\documentclass[runningheads]{llncs}

\usepackage{eccv}

\usepackage{eccvabbrv}

\usepackage{epsfig}
\usepackage{url}            %
\usepackage{booktabs}       %
\usepackage{amsfonts}       %
\usepackage{nicefrac}       %
\usepackage{microtype}      %
\usepackage{xcolor}         %
\usepackage{colortbl}         %

\usepackage{graphicx}
\usepackage{tikz}
\usepackage{amsmath}
\usepackage{amssymb}
\usepackage{bbm}
\usepackage{bm}
\usepackage{booktabs}
\usepackage{alias}
\usepackage{lib}
\usepackage{cuted}
\usepackage{xcolor}
\usepackage{array}
\usepackage{multirow}
\newcolumntype{C}[1]{>{\centering\arraybackslash}p{#1}}
\usepackage{wrapfig}
\usepackage{pifont}%

\usepackage[accsupp]{axessibility}  %

\usepackage{hyperref}

\usepackage{orcidlink}

\begin{document}

\title{3D Hand Pose Estimation in \\ Everyday Egocentric Images}

\author{Aditya Prakash \and Ruisen Tu \and Matthew Chang \and Saurabh Gupta}

\authorrunning{A.~Prakash et al.}

\institute{University of Illinois Urbana-Champaign \\
\email{\{adityap9,ruisent2,mc48,saurabhg\}@illinois.edu}\\
\url{https://bit.ly/WildHands}}

\maketitle

\definecolor{mypurple}{RGB}{0, 128, 255}
\definecolor{dexcolor}{RGB}{0, 117, 4}
\definecolor{honotatecolor}{RGB}{1, 0, 253}
\newcommand{\TODO}[1]{\textcolor{red}{#1}} %
\newcommand{\review}[1]{#1} %
\newenvironment{comments}
  {\color{red}\subsubsection*{Comments}}
  {}

\newenvironment{remove}
  {\color{red}\subsubsection*{Remove:}}
  {}

\newcommand{\xx}{\textcolor{red}{[XX]}\xspace}
\newcommand{\XX}{\xx}

\newcommand{\blue}[1]{{\color{blue}#1}}
\newcommand{\red}[1]{{\color{red}#1}}

\newcommand{\colorRef}[1]{\textcolor{red}{#1}} %
\newcommand{\reffig}[1]{\colorRef{Fig.~\ref{#1}}}
\newcommand{\reffigure}[1]{\colorRef{Figure \ref{#1}}}
\newcommand{\refFig}[1]{\mbox{\colorRef{Figure~\ref{#1}}}}
\newcommand{\reftab}[1]{\colorRef{Tab.~\ref{#1}}}
\newcommand{\refTab}[1]{\mbox{\colorRef{Table~\ref{#1}}}}
\newcommand{\refeq}[1]{\colorRef{Eq.~\ref{#1}}}
\newcommand{\refEq}[1]{\mbox{\colorRef{Equation~\ref{#1}}}}
\newcommand{\refsec}[1]{\colorRef{Sec.~\ref{#1}}}
\newcommand{\refSec}[1]{\colorRef{Section~\ref{#1}}}
\newcommand{\ccite}[1]{~\cite{#1}}

\definecolor{GreenColor}{rgb}{0.137,0.573,0.565}
\definecolor{OrangeColor}{rgb}{0.914,0.541,0.0.141}
\definecolor{PurpleColor}{rgb}{0.5,0,0.7}
\definecolor{BlueColor}{rgb}{0,0.725,0.949}
\definecolor{PinkColor}{rgb}{0.9843,0.19215,0.6}

\newcommand{\rb}{\textcolor{OrangeColor}{\textbf{R2}}\xspace}
\newcommand{\rc}{{\color{BlueColor}\textbf{R3}}\xspace}

\newcommand{\M}[1]{\mathbf{#1}} %
\newcommand{\V}[1]{\mathbf{#1}} %
\newcommand{\R}{\rm I\!R}
\newcommand{\E}{\rm I\!E}
\newcommand{\loss}{\mathcal{L}}

\newcommand{\myparagraph}[1]{\noindent\textbf{#1:}}

\newcommand{\nameCOLOR}[1]{\textcolor{black}{#1}} %

\newcommand{\datasetname}{\mbox{\nameCOLOR{ARCTIC}}\xspace}
\newcommand{\datasetfullname}{\textbf{A}\textbf{R}ticulated obje\textbf{C}\textbf{T}s in \textbf{I}ntera\textbf{C}tion}
\newcommand{\methodnameSF}{\mbox{\nameCOLOR{ArcticNet-SF}}\xspace}
\newcommand{\methodnameLSTM}{\mbox{\nameCOLOR{ArcticNet-LSTM}}\xspace}
\newcommand{\interfieldSF}{\mbox{\nameCOLOR{InterField-SF}}\xspace}
\newcommand{\interfieldLSTM}{\mbox{\nameCOLOR{InterField-LSTM}}\xspace}

\newcommand{\intermethod}{\mbox{\nameCOLOR{InterField}}\xspace}
\newcommand{\taskpose}{consistent motion reconstruction\xspace}
\newcommand{\taskfield}{interaction field estimation\xspace}
\newcommand{\taskPose}{Consistent motion reconstruction\xspace}
\newcommand{\taskField}{Interaction field estimation\xspace}

\newcommand{\highlightNUMB}[1]{\textcolor{black}{#1}} %
\newcommand{\numCleaners}{{\highlightNUMB{$4$}}\xspace}
\newcommand{\numSeqs}{{\highlightNUMB{$339$}}\xspace}
\newcommand{\numObjs}{{\highlightNUMB{$11$}}\xspace}
\newcommand{\numSubs}{{\highlightNUMB{$10$}}\xspace}
\newcommand{\numMales}{{\highlightNUMB{$5$}}\xspace}
\newcommand{\numFemales}{{\highlightNUMB{$5$}}\xspace}
\newcommand{\numAllo}{{\highlightNUMB{$8$}}\xspace}
\newcommand{\numEgo}{{\highlightNUMB{$1$}}\xspace}
\newcommand{\fps}{{\highlightNUMB{$30$}}\xspace}
\newcommand{\numImages}{{\highlightNUMB{$2.1$M}}\xspace}
\newcommand{\numVicon}{{\highlightNUMB{$54$}}\xspace}
\newcommand{\smallMarkersize}{{\highlightNUMB{$1.5$mm}}\xspace}
\newcommand{\mediumMarkersize}{{\highlightNUMB{$4.5$mm}}\xspace}

\newcommand{\video}{\textcolor{red}{\textbf{video}}\xspace}
\newcommand{\suppl}{\textcolor{black}{SupMat}\xspace}

\newcommand{\freeform}{free-form\xspace}

\newcommand{\rgbD}{\mbox{RGB-D}\xspace}

\newcommand{\oneD}{{1D}\xspace}
\newcommand{\twoD}{{2D}\xspace}
\newcommand{\sixD}{{6D}\xspace}
\newcommand{\sevenD}{{7D}\xspace}
\newcommand{\threeD}{\xspace{3D}\xspace}
\newcommand{\pytorch}{\mbox{PyTorch}\xspace}
\newcommand{\mocap}{\mbox{MoCap}\xspace}
\newcommand{\moshpp}{\mbox{MoSh++}\xspace}
\newcommand{\mosh}{\moshpp}
\newcommand{\etend}{{end-to-end}\xspace}
\newcommand{\sota}{{state-of-the-art}\xspace}
\newcommand{\inthewild}{{in-the-wild}\xspace}
\newcommand{\tpose}{\mbox{T-Pose}\xspace}
\newcommand{\blender}{{Blender}\xspace}
\newcommand{\groundtruth}{{ground-truth}\xspace}
\newcommand{\vicon}{\mbox{Vicon}\xspace}
\newcommand{\shogun}{\xspace{\mbox{Sh\={o}gun}}\xspace}
\newcommand{\shogunLive}{\mbox{\shogun-Live}\xspace}
\newcommand{\shogunPost}{\mbox{\shogun-Post}\xspace}
\newcommand{\smpl}{\mbox{SMPL}\xspace}
\newcommand{\smplx}{\mbox{SMPL-X}\xspace}
\newcommand{\smplX}{\smplx}
\newcommand{\hmr}{\mbox{HMR}\xspace}
\newcommand{\spec}{\mbox{SPEC}\xspace}
\newcommand{\mano}{\mbox{MANO}\xspace}
\newcommand{\grab}{\mbox{GRAB}\xspace}
\newcommand{\honotate}{\mbox{HO-3D}\xspace}
\newcommand{\dexycb}{\mbox{DexYCB}\xspace}
\newcommand{\tsne}{\mbox{T-SNE}\xspace}
\newcommand{\interhand}{\mbox{InterHand2.6M}\xspace}
\newcommand{\interHand}{\interhand}
\newcommand{\internet}{\mbox{InterNet}\xspace}
\newcommand{\interNet}{\internet}
\newcommand{\pointnet}{\mbox{PointNet}\xspace}

\begin{abstract}
3D hand pose estimation in everyday egocentric images is challenging
for several reasons: poor visual signal (occlusion from the object of
interaction, low resolution \& motion blur), large perspective distortion
(hands are close to the camera), and lack of 3D annotations outside of
controlled settings.
While existing methods often use hand crops as input to focus on
fine-grained visual information to deal with poor visual signal, the
challenges arising from perspective distortion and lack of 3D annotations
in the wild have not been systematically studied. We focus on this gap and
explore the impact of different practices, \ie crops as input,
incorporating camera information, auxiliary supervision, scaling up
datasets. We provide several insights that are applicable to both convolutional and transformer models, leading to better performance. Based on our findings, we also present \name, a system for 3D hand pose estimation in everyday egocentric images. Zero-shot evaluation on 4
diverse datasets (H2O, \assembly, \epic, \egoexo) demonstrate the
effectiveness of our approach across 2D and 3D metrics, where we beat past
methods by 7.4\% -- 66\%. In system level comparisons, \name achieves the best
3D hand pose on \arctic egocentric split, outperforms
FrankMocap across all metrics and HaMeR on 3 out of 6 metrics while being 10$\times$ smaller and trained on 5$\times$ less data.
\keywords{3D Hand Pose \and Egocentric Vision \and 3D from single image} %
\end{abstract}

\newcommand{\tile}[6]{
  \resizebox{0.5\linewidth}{!}{
  \begin{tikzpicture}[scale=1.00]
    \node[inner sep=0pt] (vis) at (0,-0.01\linewidth)
    {\includegraphics[width=.4\linewidth,trim=#6,clip]{#1-crop.pdf}};
    \node[inner sep=0pt] (vis) at (0.5\linewidth+13pt,0.09\linewidth)
    {\includegraphics[width=.6\linewidth]{#1-#2.pdf}};
    \node[inner sep=0pt] (vis) at (0.5\linewidth+13pt,-0.11\linewidth)
    {\includegraphics[width=.6\linewidth]{#1-#4.pdf}};
    \node[rotate=90] (first) at (0.2\linewidth+7pt, 0.09\linewidth){\small #3};
    \node[rotate=90] (first) at (0.2\linewidth+7pt, -0.11\linewidth){\small #5};
  \end{tikzpicture}
  }
}

\begin{figure}[t]
  \tile{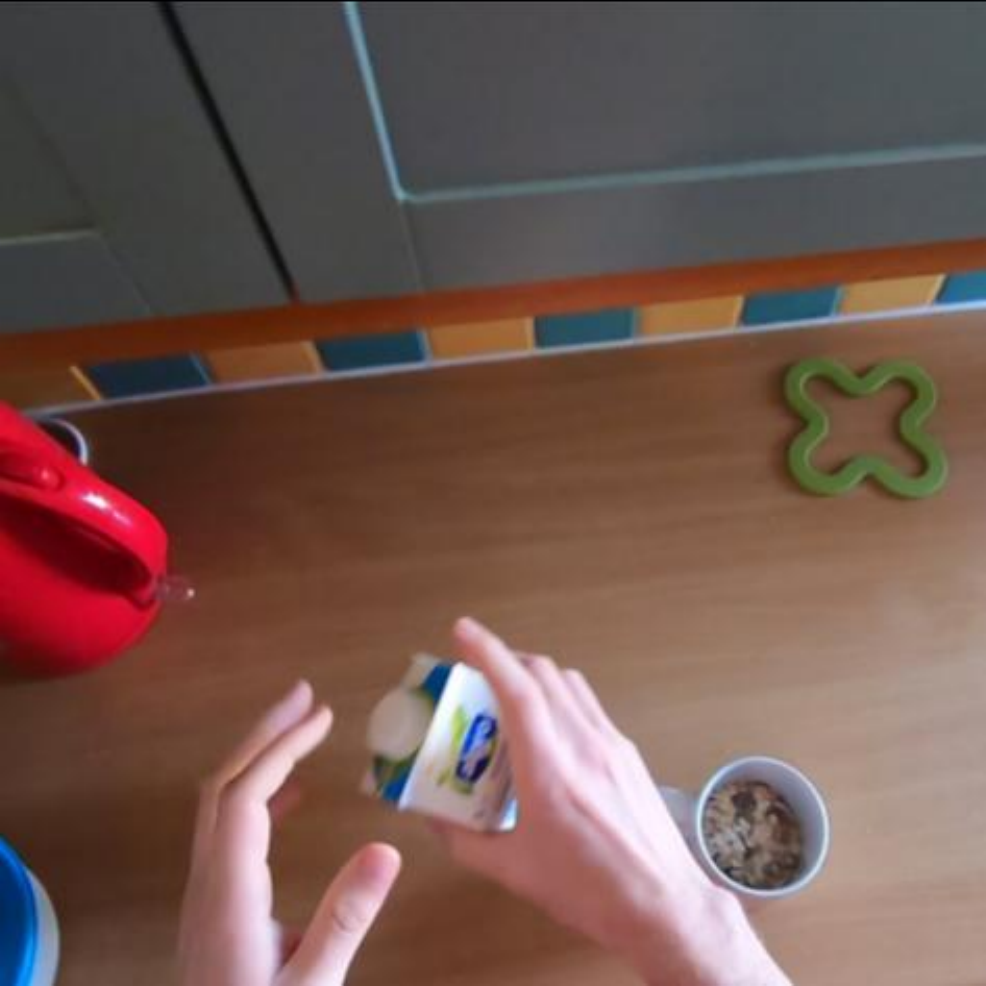}{fm}{FrankMocap}{wh}{WildHands}{0 0 0 0}
  \tile{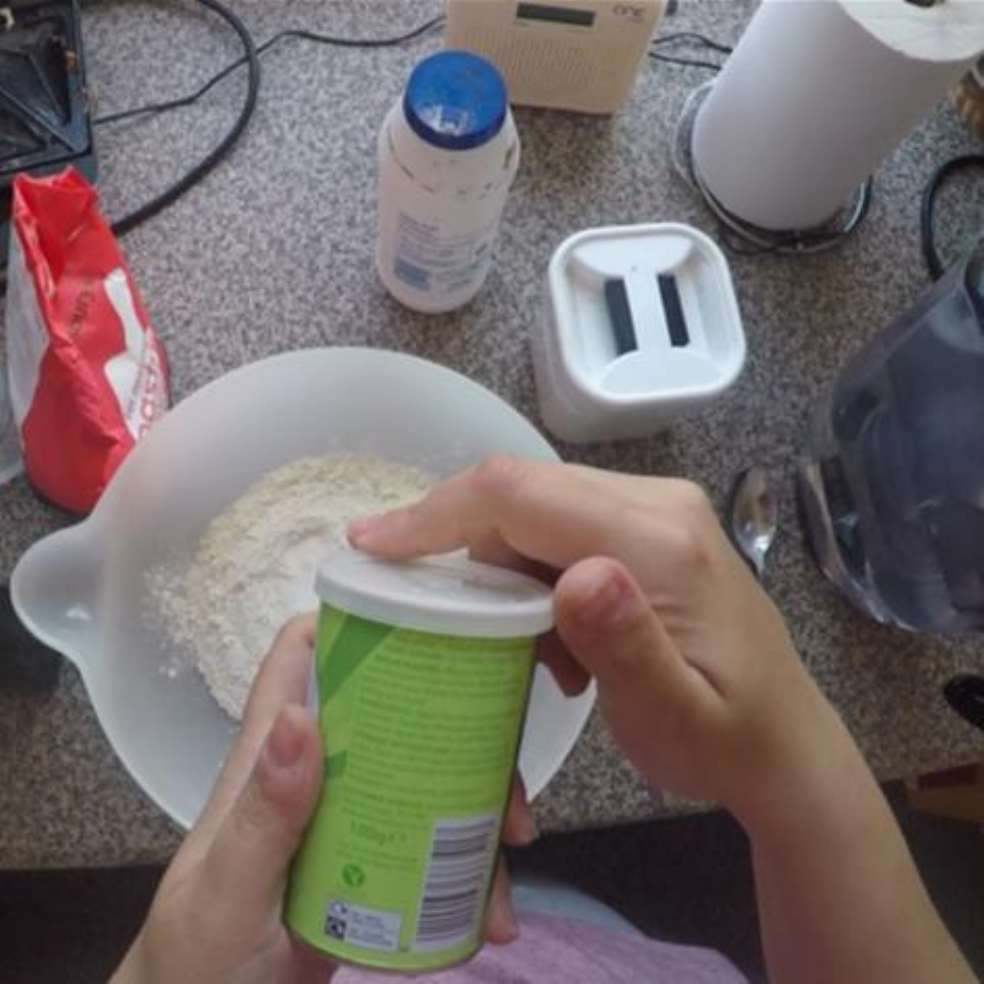}{fm}{FrankMocap}{wh}{WildHands}{0 0 0 0}\\
  \tile{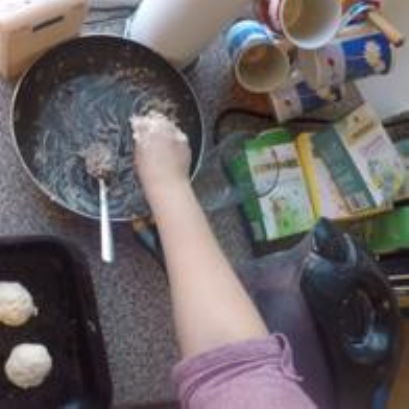}{hm}{HaMeR}{wh}{WildHands}{0 0 0 0}
  \tile{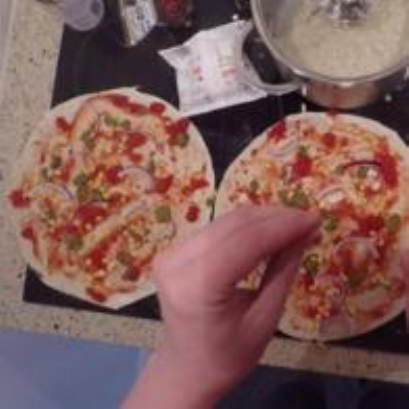}{hm}{HaMeR}{wh}{WildHands}{0 0 0 0}
\caption{\textbf{\name} predicts the 3D shape, 3D articulation and 3D placement of the hand in the camera frame from a single in-the-wild egocentric \rgb image and camera intrinsics. It produces better 3D output compared to FrankMocap~\cite{rong2020frankmocap} in occlusion scenarios and is more adept at dealing with perspective distortion than HaMeR~\cite{pavlakos2023reconstructing}, in challenging egocentric hand-object interactions from \epic~\cite{damen2018scaling} dataset.}
\figlabel{teaser}
\end{figure}

\section{Introduction}
\seclabel{intro}

Understanding egocentric hands in 3D enables applications in AR/VR, robotics.  While several works have studied exocentric hands~\cite{rong2020frankmocap,
pavlakos2023reconstructing}, no existing approach performs well in diverse egocentric settings outside of lab setups. We focus on this gap \& study the impact of common practices, \ie crops as input, camera information, auxiliary supervision, scaling up datasets, for predicting absolute 3D hand pose from a single egocentric image. We identify 2 important factors: a) modeling the 3D to 2D projection during imaging of the hand in egocentric
views, b) scaling up training to diverse datasets by leveraging auxiliary supervision.

Let's unpack each component. Existing methods often operate
on image crops, assume that the image crop is located at the center of the camera's field of view with a made-up focal length. These choices are reasonable for exocentric settings where the location of the hand in the image does not provide any signal for the hand articulation; and perspective distortion effects are minimal as the hand is far away \& occupies a relatively small part of the camera's field of view.  However, these assumptions are sub-optimal for processing egocentric images.

Due to the biomechanics of the hand, its location in egocentric images carries information about its pose. Also, as the hand is closer to the camera in egocentric settings, it undergoes a lot more perspective distortion than in exocentric images. 3D hand pose that correctly explains the 2D hand appearance in one part of an egocentric image, may not be accurate for another part of the image. Thus, the location of the hand in the image must be taken into account while making 3D predictions. This suggests feeding the 2D location of the hand in the image to the network. However, the notion of 2D location in the image frame is camera specific. The more fundamental quantity that generalizes across cameras, is the {\it angular location in the camera's field of view}. We thus adopt the recent KPE embedding~\cite{Prakash2023Ambiguity} to augment hand crop features with sinusoidal encodings of its location in the camera's field of view \& find this to improve performance.

However, just processing image crops the right way is not sufficient for generalization. The model also needs to be trained on broad \& diverse datasets outside of lab settings. This is not easy as 3D hand pose is difficult to directly annotate in images. We thus turn to joint training on 3D
supervision from lab datasets and 2D auxiliary supervision on in-the-wild data in the form of 2D hand  masks~\cite{darkhalil2022visor,cheng2023towards} \& grasp labels~\cite{cheng2023towards}. To absorb supervision from segmentation labels, we differentiably render~\cite{Liu2019ICCV} the predicted 3D hand into images and back-propagate the loss through the rendering. For grasp supervision, we note that hand  pose is indicative of the grasp type and use supervision from a grasp classifier that takes the predicted 3D hand pose as input.

Lack of accurate 3D annotations outside of lab settings makes it challenging to assess the generalization capabilities. To this end, we adopt a {\it zero-shot} evaluation strategy. Even though a single lab dataset has limited diversity, a model that performs well on a lab dataset {\it without having seen any images from it} likely generalizes well. Furthermore, we collect \epichands, containing 2D hand joint annotations on 5K images from the \visor~\cite{darkhalil2022visor} split of in-the-wild \epic~\cite{Damen2018ScalingEV} to evaluate the 2D projections of the predicted 3D hand pose on everyday images. We also consider the 3D hand poses provided evaluate on the concurrent \egoexo~\cite{grauman2023ego}. We believe that these evaluations together comprehensively test the generalization capabilities of different models.

Our experiments (\Secref{sec:experiments}) show the utility of (1) using crops (\vs full images), (2) inputting 2D crop
location (\vs not), (3) encoding the crop's location in camera's field of view (\vs in the image frame), and (4) 2D mask \& grasp supervision. We apply these insights to both convolutional and transformer models, leading to better performance. We also present \name (\Figref{teaser}) which outperforms FrankMocap~\cite{rong2020frankmocap} on egocentric images and is competitive to concurrent HaMeR~\cite{pavlakos2023reconstructing} while being $10\times$ smaller \& trained with $5\times$ less data.

\section{Related Work}
\label{sec:related}

\boldparagraph{Hand pose estimation \& reconstruction} Several decades of
work~\cite{rehg1994visual, heap1996towards, freeman1995orientation} have
studied different aspects: 2D pose~\cite{Simon2017CVPR,Chao2021CVPR} \vs 3D
pose~\cite{tompson2014real,ohkawa2023assemblyhands,kwon2021h2o,Wan2016ECCV} \vs
mesh~\cite{Ballan2012ECCV,Sridhar2013ICCV,hasson19_obman},
RGB~\cite{hasson19_obman,Hampali2022CVPR,Fan2023CVPR} \vs
RGBD~\cite{Simon2017CVPR,Sun2015CVPR,rogez20143d, tompson2014real,
sharp2015accurate, sridhar2016real} inputs,
egocentric~\cite{Fan2023CVPR,ohkawa2023assemblyhands} \vs
allocentric~\cite{hampali2020honnotate,Hampali2022CVPR,Fan2023CVPR}, hands in
isolation~\cite{Moon2020ECCV,Zimmermann2019ICCV} \vs interaction with
objects~\cite{hampali2020honnotate,Liu2022CVPR,Yang2022CVPR}, feed-forward
prediction~\cite{hasson19_obman,Sener2022CVPR,Hampali2022CVPR,Fan2023CVPR} \vs
test-time optimization~\cite{cao2021reconstructing,Hasson2020LeveragingPC}.
Driven by the advances in parametric hand models~\cite{romero2017embodied,
Potamias_2023_CVPR}, recent work has moved past 3D joint estimation towards 3D
mesh recovery~\cite{zhang2019end, hasson19_obman, rong2020frankmocap,
Fan2023CVPR, Hampali2022CVPR, pavlakos2023reconstructing} in 3
contexts: single hands in isolation~\cite{zimmermann2017learning}, hands
interacting with objects~\cite{Fan2023CVPR, tzionas20153d} and two hands
interacting with one another~\cite{Moon2020ECCV, Hampali2022CVPR}. Jointly
reasoning about hands \& objects has proved fruitful to improve both hand \&
object reconstruction~\cite{ye2022ihoi, hasson19_obman,
karunratanakul2020grasping}. While several expressive models focus on 3D hand pose estimation in lab
settings~\cite{Hampali2022CVPR,Sener2022CVPR,Jiang2023CVPR,JiangR2023CVPRa,Ivashechkin2023ARXIV}, only a very few works~\cite{pavlakos2023reconstructing} tackle the problem in
everyday egocentric images as in Ego4D~\cite{grauman2022ego4d}, Epic-Kitchen~\cite{Damen2018ScalingEV}. We focus on this setting
due to challenges involving perspective distortion, dynamic interactions \& heavy occlusions. We explore both convolutional~\cite{rong2020frankmocap,Fan2023CVPR} and transformer models~\cite{Park2022CVPR,pavlakos2023reconstructing} to study the impact of using crops, location of the crop in camera's field of view \& auxiliary supervison in zero-shot generalization to diverse egocentric settings.

\boldparagraph{Hand datasets} Since 3D hand annotations from single images is difficult to get, most datasets are collected in controlled settings to get 3D ground truth using MoCap~\cite{Fan2023CVPR, taheri2020grab}, multi-camera setups~\cite{ohkawa2023assemblyhands, kwon2021h2o,hampali2020honnotate,Hampali2022CVPR,Liu2022CVPR}, or magnetic sensors~\cite{garcia2018first}. They often include single hands in isolation~\cite{Zimmermann2019ICCV}, hand-object interactions~\cite{Fan2023CVPR,hampali2020honnotate,kwon2021h2o,Hampali2022CVPR} \& hand-hand interactions~\cite{Moon2020ECCV}. Different from these datasets with 3D poses, ~\cite{darkhalil2022visor, shan2020understanding, cheng2023towards} provide annotations for segmentation masks~\cite{darkhalil2022visor,cheng2023towards}, 2D bounding boxes~\cite{shan2020understanding} and grasp labels~\cite{cheng2023towards} on internet videos~\cite{shan2020understanding} and egocentric images in the wild~\cite{damen2018scaling,grauman2022ego4d}. Our work combines 3D supervision from datasets~\cite{Fan2023CVPR,ohkawa2023assemblyhands} captured in controlled settings with
2D auxiliary supervision, \ie segmentation masks \& grasp labels, from datasets
outside the lab~\cite{darkhalil2022visor, cheng2023towards} to learn models that perform
well in challenging everyday images. We collect \epichands dataset with 2D
hand keypoints on 5K images from \epic for evaluation in everyday images
outside of lab settings. We also use concurrent \egoexo~\cite{grauman2023ego} that annotates 2D keypoints in paired ego \& exo views to get 3D hand annotations.

\begin{figure*}[t]
    \centering
    \includegraphics[width=1.0\textwidth]{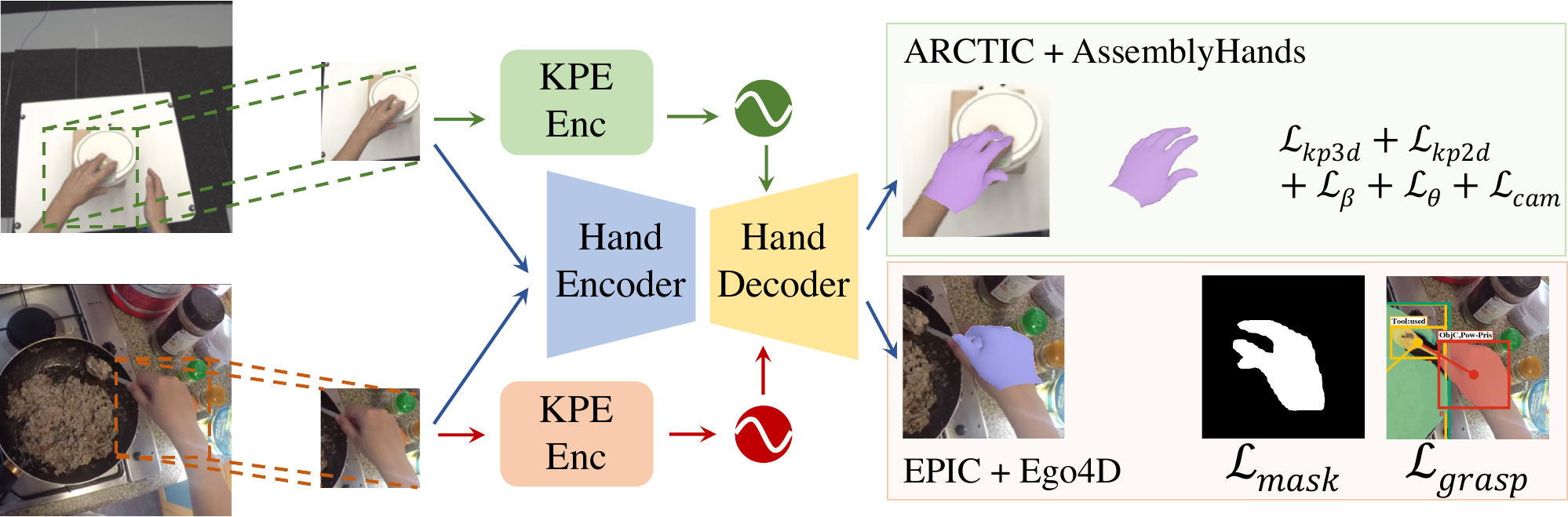}
    \caption{\textbf{Model Overview}. We crop the input images around the hand
    and process them using a convolutional backbone. The hand features along
    with the global image features (not shown above for clarity) and
    intrinsics-aware positional encoding (\camenc~\cite{Prakash2023Ambiguity})
    for each crop are fed to the decoder to predict the 3D hand. The hand
      decoders predict MANO parameters $\beta, \art, \pose$ and camera
      translation which are converted to 3D keypoints \& 2D keypoints and
      trained using 3D supervision on lab datasets, \eg
      ARCTIC~\cite{Fan2023CVPR}, AssemblyHands~\cite{ohkawa2023assemblyhands}.
      We also use auxiliary supervision from in-the-wild
      \epic~\cite{darkhalil2022visor} dataset via hand segmentation masks and
      grasp labels. The hand masks are available with the VISOR
      dataset~\cite{darkhalil2022visor} whereas grasp labels are estimated
      using off-the-shelf model from~\cite{cheng2023towards}.}
    \figlabel{model}
\end{figure*}

\boldparagraph{Auxiliary supervision} Several works on 3D shape prediction from a single image~\cite{kanazawa2018end,tulsiani2017multi} often use auxiliary supervision to deal with lack of 3D annotations. \cite{kanazawa2018end} uses keypoint supervision for 3D human mesh recovery, while \cite{tulsiani2017multi} uses multi-view consistency cues for 3D object reconstruction. Aided by differentiable rendering~\cite{kato2018neural, liu2020general}, segmentation and depth prediction have been used to provide supervision for 3D reconstruction~\cite{kanazawa2018learning, cao2021reconstructing, Hasson2020LeveragingPC}. We adopt this use of segmentation as an auxiliary cue for 3D poses. In addition, we use supervision from hand grasp labels based on the insight that hand grasp is indicative of the hand pose.

\boldparagraph{Ambiguity}
3D estimation from a single image is ill-posed due to ambiguities arising from scale-depth confusion~\cite{hartley2003multiple} and cropping~\cite{Prakash2023Ambiguity}. Recent work~\cite{Prakash2023Ambiguity} points out the presence of perspective distortion-induced shape ambiguity in image crops and uses camera intrinsic-based location encodings to mitigate it. We investigate the presence of this ambiguity for hand crops in egocentric images and adopt the proposed embedding to mitigate it. Similar embeddings have been used before in literature, primarily from the point of view of training models on images from different cameras~\cite{facil2019cam, guizilini2023towards}, to encode extrinsic information~\cite{guizilini2022depth, yifan2022input,miyato2023gta}.

\section{Method}
\seclabel{method}

We present \name, a new system for 3D hand pose estimation from egocentric images in the wild. We build on top of ArcticNet-SF~\cite{Fan2023CVPR} and FrankMocap~\cite{rong2020frankmocap}. Given a crop around a hand and associated camera intrinsics, \name predicts the 3D hand shape as MANO~\cite{romero2017embodied} parameters, shape $\beta$ and pose $\theta$. $\theta$ consists of angles of articulation $\art$ for 15 hand joints and the global pose $\pose$ of the root joint in the camera coordinate system. \name is trained using both lab (\arctic, \assembly) and in-the-wild (\epicb, \ego) datasets with different sources of supervision. \Figref{model} provides an overview of our model. Next, we describe each component of \name in detail.

\subsection{Architecture}
\seclabel{sec:architecture}

\boldparagraph{Hand encoder}
Our models uses hand crops as input (resized to $224 \times 224$ resolution),
which are processed by a ResNet50~\cite{He2016CVPR} backbone to get $7 \times 7
\times 2048$ feature maps. The left and right hand crops are processed separately but the parameters are shared. We also use global image features in our model, computed by average pooling the $7 \times 7 \times 2048$ feature map to get a $2048$-dimensional vector.

\boldparagraph{Incorporating \camenc}
Recent work~\cite{Prakash2023Ambiguity} has shown that estimating 3D quantities from image crops suffers from perspective distortion-induced shape ambiguity~\cite{Prakash2023Ambiguity}. This raises concerns about whether this ambiguity is also present when using hand crops for predicting 3D pose and how to deal with it. Following the study in~\cite{Prakash2023Ambiguity}, we analyze the hands in the \arctic dataset (details in the supplementary) and find evidence of this ambiguity in hand crops as well. Thus, we adopt the intrinsics-aware positional encoding (\camenc) proposed in~\cite{Prakash2023Ambiguity} to mitigate this ambiguity. Specifically, we provide the network with information about the location of the hand crop in the field of view of the camera. Consider the principal point as $(p_x, p_y)$ \& focal length as $(f_x, f_y)$. For each pixel $(x,y)$, we compute $\theta_x = \tan^{-1} \left( \frac {x - p_x}{f_x} \right)$, $\theta_y = \tan^{-1} \left( \frac {y - p_y}{f_y} \right)$ \& convert them into sinusoidal encoding~\cite{Mildenhall2020ECCV}.

We add \camenc to the $7 \times 7 \times 2048$ feature map. \camenc comprises sinusoidal encoding of the angles $\theta_x$ and $\theta_y$ (Sec. 4.1 in the main paper), resulting in $5*4*K$ dimensional sparse encoding (4 for corners and 1 for center pixel) and $H \times W \times 4*K$ resolution dense encoding, where $K$ is the number of frequency components (set to 4). For the sparse \camenc variant, we broadcast it to $7 \times 7$ resolution whereas for the dense \camenc variant, we interpolate it to $7 \times 7$ resolution and concatenate to the feature map. This concatenated feature is passed to a 3 convolutional layers (with 1024, 512, 256 channels respectively, each with kernel size of $3 \times 3$ and ReLU~\cite{Nair2010ICML} non-linearity) to get a $3 \times 3 \times 256$ feature map. This is flattened to 2304-dimensional vector and passed through a 1-layer MLP to get a 2048-dimensional feature vector. We do not use batchnorm~\cite{Ioffe2015ICML} here since we want to preserve the spatial information in \camenc. %

\boldparagraph{Hand decoder}
It consists of an iterative architecture, similar to decoder in HMR~\cite{kanazawa2018end}. The inputs are the $2048$-dimensional feature vector and initial MANO~\cite{romero2017embodied} (shape $\beta$, articulation $\art$ and global pose $\pose$, all initialized as 0-vectors) \& weak perspective camera parameters (initialized from the $2048$-dimensional feature vector). Each of these parameters are predicted using a separate decoder head. The rotation parameters $\art$, $\pose$ are predicted in matrix form and converted to axis-angle representation to feed to MANO model. Each decoder is a 3-layer MLP with the 2 intermediate layers having 1024 channels and the output layer having the same number of channels as the predicted parameter. The output of each decoder is added to the initial parameters to get the updated parameters. This process is repeated for 3 iterations. The output of the last iteration is used for the final prediction.

\boldparagraph{Differentiable rendering for mask prediction}
The outputs from the decoder, $\beta$, $\art$, and $\pose$ for the predicted
hand, are passed to a differentiable MANO
layer~\cite{romero2017embodied,hasson19_obman} to get the hand mesh. This is
used to differentiably render a soft segmentation mask, $M$, using
SoftRasterizer~\cite{liu2020general,ravi2020pytorch3d}. Using a differentiable
hand model (MANO) and differentiable rendering lets us train our model end-to-end.

\boldparagraph{Grasp classifier}
We use the insight that grasp type during interaction with objects is
indicative of hand pose. We train a grasp prediction head on $\art$, $\pose$ \&
$\beta$ (predicted by \name) via a 4-layer MLP (with 1024, 1024, 512, 128 nodes
\& ReLU non-linearity after each). The MLP predicts logits for the 8 grasp
classes defined in~\cite{cheng2023towards} which are converted into
probabilities, $G$ via softmax.

\subsection{Training supervision}
We train \name using: (1) 3D supervision on $\beta$, $\art$, $\pose$, 3D hand keypoints \& 2D projections of 3D keypoints in the image on lab datasets, and (2) hand masks and grasp labels on in-the-wild datasets. %
\begin{align}
    &\mathcal{L}_{\theta} = {\lVert \theta-\theta^{gt} \rVert}_2^2 \qquad
    \mathcal{L}_{\beta} = {\lVert \beta-\beta^{gt} \rVert}_2^2 \qquad
    \mathcal{L}_{cam} = {\lVert (s, T)-(s, T)^{gt} \rVert}_2^2 \\
    &\mathcal{L}_{kp3d} ={\lVert J_{3D}-J_{3D}^{gt} \rVert}_2^2 \qquad
    \mathcal{L}_{kp2d} ={\lVert J_{2D} - J_{2D}^{gt} \rVert}_2^2 \\
    &\mathcal{L}_{mask} = {\lVert M-M^{gt} \rVert} \qquad
    \mathcal{L}_{grasp} = CE(G, G^{gt})
\end{align}
Here, $\mathcal{L}_{\theta}$ is used for both $\theta_{local}$ \&
$\theta_{global}$, $(s,T)$ are the weak perspective camera parameters and $CE$
represents cross-entropy loss. $J_{2D} = K[J_{3D} + (T, f/s)]$, where $J_{3D}$
is the 3D hand keypoints in the MANO coordinate frame, $K$ is the camera
intrinsics, $f$ is the focal length, and $s$ is the scale factor of the weak
perspective camera. Note that $(\onedot)^{gt}$ represents the ground truth
quantities. The total loss is:
\begin{align}
    \mathcal{L} & = \lambda_{\theta} \mathcal{L}_{\theta} + \lambda_{\beta}
    \mathcal{L}_{\beta} + \lambda_{cam} \mathcal{L}_{cam} 
    + \lambda_{kp3d} \mathcal{L}_{kp3d} + \lambda_{kp2d} \mathcal{L}_{kp2d} \nonumber\\
    & + \lambda_{mask} \mathcal{L}_{mask} + \lambda_{grasp} \mathcal{L}_{grasp}
\end{align}
\boldparagraph{Lab datasets} For \arctic, we use $\lambda_{\theta}=10.0, \lambda_{\beta}=0.001, \lambda_{kp3d}=5.0, \lambda_{kp2d}=5.0, \mathcal{L}_{cam}=1.0$ \& set other loss weights to 0. \assembly does not use MANO representation for hands, instead provides labels for 3D \& 2D keypoints of 21 hand joints. So, we use $\lambda_{kp3d}=5, \lambda_{kp2d}=5$ \& set other loss weights to 0.

\boldparagraph{In-the-wild data}
For \epicb \& \ego, we use hand masks \& grasp labels as auxiliary supervision. While \visor contains hand masks, grasp labels are not available. \ego does not contain either hand masks or grasp labels. To extract these labels, we use predictions from off-the-shelf model~\cite{cheng2023towards} as pseudo ground truth. We use $\lambda_{mask}=10.0, \lambda_{grasp}=0.1$ \& set other loss weights to 0.

\subsection{Implementation Details}
\seclabel{details}

Our model takes hand crops as input. During training, we use the ground truth bounding box for the hand crop (with small perturbation), estimated using the 2D keypoints \& scaled by a fixed value of 1.5 to provide additional context around the hand. At test time, we need to predict the bounding box of the hand in the image. On \arctic, we train a bounding box predictor on by finetuning MaskRCNN~\cite{He2017ICCV}. This is also used for submitting the model to the \arctic leaderboard. For \epichands, we use the recently released hand detector from~\cite{chen2019learning}. All the ablations use ground truth bounding box for the hand crop.

We use the training sets of \arctic (187K images) \& \assembly (360K), \visor split (30K) of \epicc and 45K images from \ego kitchen videos to train our model. \name is trained jointly on different datasets with the input batch containing images from multiple datasets. All models are initialized from the ArcticNet-SF model trained on the allocentric split of the ARCTIC dataset~\cite{Fan2023CVPR}. All models are trained for 100 epochs with a learning rate of $1e-5$. The multi-dataset training is done on 2 A40 GPUs with a batch size of 144 and Adam optimizer~\cite{Kingma2015ICLR}. More details are provided in the supplementary.

\section{Experiments}
\seclabel{sec:experiments}

We adopt a zero-shot evaluation strategy: 3D evaluation on lab datasets (H2O, \assembly), evaluation of 2D projections of 3D hand predictions on \epichands \& 3D evaluation on EgoExo4D~\cite{grauman2023ego}. We systematically analyze the
effectiveness of design choices (using crops, \camenc), different terms in the loss function and different datasets used for training. We also report a system-level comparison on \arctic leaderboard and with FrankMocap~\cite{rong2020frankmocap} \&
HaMeR~\cite{pavlakos2023reconstructing}.

\subsection{Protocols}
\seclabel{sec:protocols}

\boldparagraph{Training datasets}
We consider 4 datasets for training: 2 lab datasets (\arctic \& \assembly) and
2 in-the-wild datasets (\epicb \& \ego).

We select \arctic since it contains the largest range of hand pose
variation~\cite{Fan2023CVPR} among existing
datasets~\cite{hampali2020honnotate,taheri2020grab,Hampali2022CVPR,Liu2022CVPR,Chao2021CVPR}.
We use the egocentric split with more than 187K images in the train set. We
also use \assembly since it is a large-scale dataset with more than 360K
egocentric images in the train split. Different combinations of these datasets
are used for different experiments.

We use egocentric images from \epic \& \ego as in-the-wild data for training
our model using auxiliary supervision. We use 30K training images available in
the \visor split of \epic and ~45K images from \ego. To extract hand masks and
grasp labels, we use off-the-shelf model from~\cite{cheng2023towards}.

\boldparagraph{Evaluation datasets}
We consider 4 datasets for zero-shot generalization experiments:
H2O~\cite{kwon2021h2o}, \assembly, \epichands, and \egoexo. Note that these
datasets cover large variation in inputs, H2O contains RGB images in lab
settings, \assembly consists of grayscale images and \epichands and \egoexo
images show hands performing everyday activities in the wild. 

We use the validation splits of H2O and \assembly with 29K and 32K images
respectively. Since 3D hand annotations are difficult to collect for
in-the-wild images, we instead collect 2D hand keypoints annotations on 5K
egocentric images from validation set of \visor split of \epicb. We refer to
this dataset as \epichands. See sample images from the dataset in
\Figref{epichands}. We also evaluate on the validation split of \egoexo hand pose dataset.

\begin{figure}[t]
    \centering
    \includegraphics[width=1.0\columnwidth]{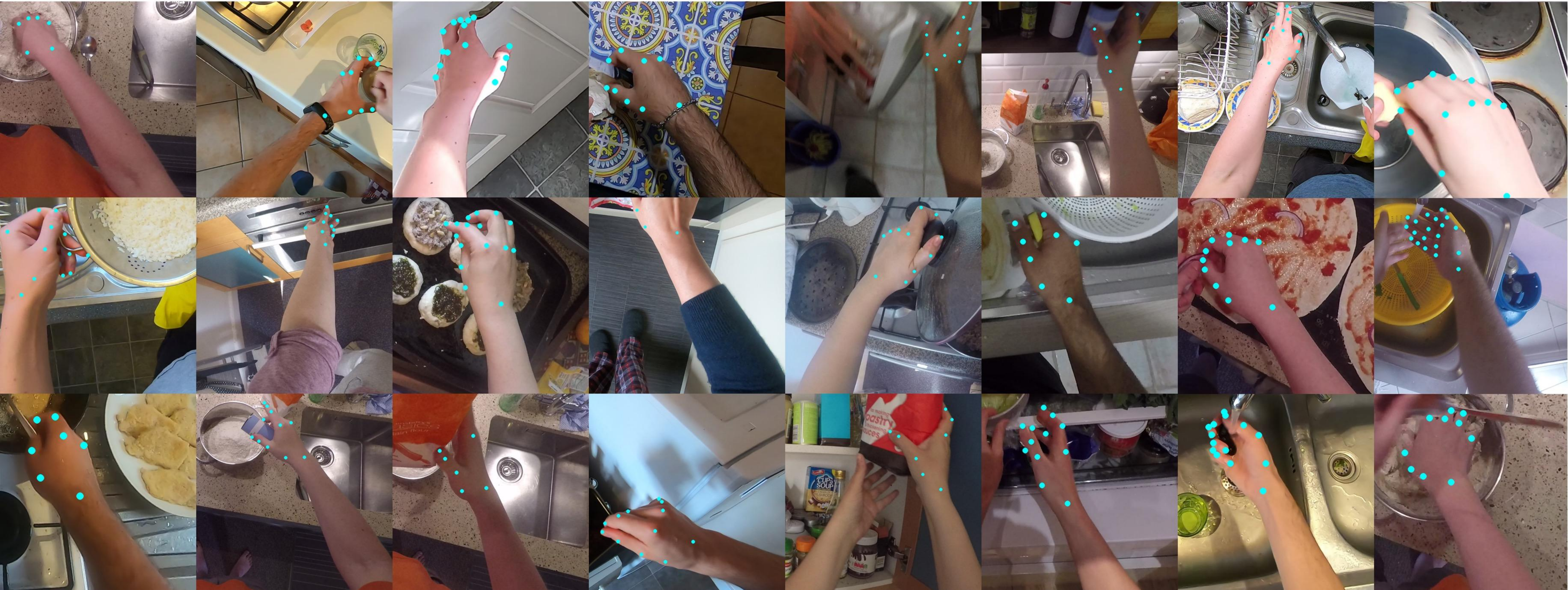}
    \caption{{\bf \epichands annotations}. We collect 2D joint annotations (shown in blue) for 5K in-the-wild egocentric images from \epic~\cite{damen2020collection}. We show few annotations here with images cropped around the hand. We also have the label for the joint corresponding to each keypoint. Note the heavy occlusion \& large variation in dexterous poses of hands interactiong with objects. More visualizations in supplementary.}
    \figlabel{epichands}
\end{figure}

\boldparagraph{\epichands}
\epichands contains 2D annotations for the 21 hand joints to facilitate
evaluation of 2D projections of the predicted 3D keypoints. We sample 5K images
from the validation set of \visor split of \epic and get the 21 joints
annotated via Scale AI. We use the same joint convention as
\arctic~\cite{Fan2023CVPR}. We crop the images around the hand using the
segmentation masks in \visor and provide the crops to annotators for labeling.
Note that most of these images do not have all the 21 keypoints visible.
Following \arctic, we only consider images with atleast 3 visible joints for
evaluation. Moreover, since the models in our experiments required hand crops
as input, we only evaluate on those images for which hand bounding box is
predicted by the recently released hand detector model
from~\cite{cheng2023towards}. This leaves us with 4724 hand annotations, with
2697 right hands and 2027 left hands. We show some annotations
in~\Figref{epichands}.

\boldparagraph{Metrics}
For 3D hand pose evaluation, we consider 2 metrics: (1) {\bf
Mean Per-Joint Position Error (MPJPE)}: L2 distance (mm) between the 21
predicted \& ground truth joints for each hand after subtracting the root
joint (this captures the relative pose). (2) {\bf Mean
Relative-Root Position Error (MRRPE)}: the metric distance between the
root joints of left hand and right hand,
following~\cite{Fan20213DV,Moon2020ECCV,Fan2023CVPR} (this takes the absolute
pose into account). 
(3) For 2D evaluation on \epichands, we measure
the {\bf L2 Error} (in pixels for 224x224 image input) between ground truth keypoints \& 2D projections of predicted 3D keypoints.

\boldparagraph{Baselines}
(1) ArcticNet-SF~\cite{Fan2023CVPR} is the single-image model released with the \arctic benchmark. It consists of a convolutional backbone (ResNet50~\cite{He2016CVPR}) to process the input image, followed by a HMR~\cite{kanazawa2018learning}-style decoder to predict the hand and object poses. The predicted hand is represented using MANO~\cite{romero2017embodied} parameterization. (2) FrankMocap~\cite{rong2020frankmocap} is trained on multiple datasets collected in controlled settings and is a popular choice to apply in in-the-wild setting~\cite{ye2022ihoi,Hasson2020LeveragingPC,cao2021reconstructing}. It uses hand crops as input instead of the entire image, which is then processed by a convolutional backbone. The decoder is similar to HMR~\cite{kanazawa2018learning} which outputs MANO parameters for hand and training is done using 3D pose \& 2D keypoints supervision. (3) HandNet: Since the training code is not available for FrankMocap, we are unable to train it in our setting. So, we implement a version of ArcticNet-SF which uses crops as input along with HMR-style decoder and train it in our setting using 3D \& 2D supervision. This baseline is equivalent to \name without \camenc and ArcticNet-SF with crops. (4) HandOccNet~\cite{Park2022CVPR}: It takes crops as input and encodes them using a FPN~\cite{Lin2017CVPR} backbone. These are passed to transformer~\cite{Vaswani2017AttentionIA} modules to get a heatmap-based intermediate representation which is then decoded to MANO parameters. (5) HaMeR~\cite{pavlakos2023reconstructing}: It also takes crops as input and processes them using a ViT~\cite{dosovitskiy2020image} backbone. The features are then passed to a transformer decoder to predict the MANO parameters. Note that adversarial loss is not used for training any model in our setting.

\begin{table*}[t]
    \centering
    \small
    \caption{{\bf Benefits of using crops and KPE.} Zero shot
    generalization performance improves through the use of crops as input
    (HandNet uses crops \vs ArcticNet-SF uses full image) and KPE helps
    (\name uses KPE with crops \vs HandNet only uses crops). All models use
    the same backbone and are trained on the same data in each setting for
    fair comparisons. $\mathcal{D}: \{ \text{\arctic, \assembly, \epicc}
    \}$. 
    }
    \tablelabel{tab:model_comparison}
    \setlength{\tabcolsep}{6pt}
    \resizebox{\linewidth}{!}{
    \begin{tabular}{l c c c c c c}
        \toprule
        & \multicolumn{2}{c}{\bf H2O} & \multicolumn{2}{c}{\bf \assemblya} & \bf \egoexo & \bf \epichands \\
        \cmidrule(lr){2-3} \cmidrule(lr){4-5} \cmidrule(lr){6-6} \cmidrule(lr){7-7}
                                             & \bf MPJPE & \bf MRRPE & \bf MPJPE & \bf MRRPE & \bf MPJPE & \bf L2 Error\\
        \midrule
        Training data & \multicolumn{2}{c}{$\mathcal{D}$} & \multicolumn{2}{c}{$\mathcal{D}$ - {\assemblya}} & $\mathcal{D}$ & $\mathcal{D}$ - {\epicc} \\
        \midrule
        ArcticNet-SF & 83.84     & 325.55    & 110.76    & 326.94     & 114.24  & 35.02 \\
        HandNet      & 38.06     & 141.06    & 109.88    & 317.49     & 89.72 & 31.62 \\
        WildHands    & \bf 31.08 & \bf 49.49 & \bf 84.91 & \bf 164.90 & \bf 55.84 & \bf 11.05 \\
        \bottomrule
    \end{tabular}
    }
\end{table*}

\begin{table*}[t]
    \centering
    \small
    \caption{{\bf Impact on transformer models.} We investigate if our insights are useful for transformer models as well, \ie if \camenc helps on top of positional encodings used in transformers \& if auxiliary supervision leads to better generalization for large capacity models. All models are trained on the same data in each setting for
    fair comparisons.
    }
    \tablelabel{tab:kpe_transformer}
    \setlength{\tabcolsep}{6pt}
    \resizebox{\linewidth}{!}{
    \begin{tabular}{l c c c c c c}
        \toprule
        & \multicolumn{2}{c}{\bf H2O} & \multicolumn{2}{c}{\bf \assemblya} & \bf \egoexo & \bf \epichands \\
        \cmidrule(lr){2-3} \cmidrule(lr){4-5} \cmidrule(lr){6-6} \cmidrule(lr){7-7}
                                             & \bf MPJPE & \bf MRRPE & \bf MPJPE & \bf MRRPE & \bf MPJPE & \bf L2 Error\\
        \midrule
        Training data & \multicolumn{2}{c}{$\mathcal{D}$} & \multicolumn{2}{c}{$\mathcal{D}$ - {\assemblya}} & $\mathcal{D}$ & $\mathcal{D}$ - {\epicc} \\
        \midrule
        HandOccNet~\cite{Park2022CVPR} & 60.58 & 187.24 & 110.28 & 293.92 & 80.96 & 32.77 \\
        HandOccNet + KPE & 47.57 & 72.25 & 103.30 & 232.83 & 78.64 & 13.54 \\
        HaMeR~\cite{pavlakos2023reconstructing} (ViT) & 30.57 & 113.26 & 79.48 & 227.59 & 55.36 & 25.48 \\
        HaMeR (ViT) + KPE & {\bf24.15} & {\bf62.99} & {\bf71.64} & {\bf184.55} & {\bf47.02} & {\bf9.77} \\
        \bottomrule
    \end{tabular}
    }
\end{table*}

\subsection{Results}
\label{sec:systematic}

We systematically study the impact of several factors: use of crops
(\tableref{tab:model_comparison}) \& \camenc
(\tableref{tab:model_comparison}, \tableref{tab:design_choices}), perspective distortion(\tableref{tab:perspective}), auxiliary supervision
(\tableref{tab:loss_ablation}), training datasets
(\tableref{tab:dataset_ablation}) on both convolutional (\tableref{tab:model_comparison}) \& transformer models (\tableref{tab:kpe_transformer}) through {\it
controlled experiments}, \ie all factors outside of what we want to check the affect of, are kept constant. All the results are reported in a {\it zero-shot setting} \ie models are not trained on the evaluation dataset.

\boldparagraph{Impact of crops} To understand the benefits due to using crops
as input instead of full images, we compare ArcticNet-SF and HandNet
in~\Tableref{tab:model_comparison}. The only difference between these two
models is: ArcticNet-SF uses full image as input whereas HandNet uses crops as
input. We see gains of 27.7\% in MPJPE, 29.7\% in MRRPE, 10.7\% in PA-MPJPE,
and 9.7\% in 2D pose across different settings. This provides evidence for the
utility of using crops as inputs~\cite{rong2020frankmocap,ohkawa2023assemblyhands}.

\boldparagraph{Benefits of \camenc} In~\Tableref{tab:model_comparison}, HandNet
\& \name differ only in the use of \camenc. This leads go improvements of
20.5\% in MPJPE, 56.4\% in MRRPE \& 65.1\% in 2D pose. Compared to impact of
crops, the gains are significantly higher in MRRPE (indicating better absolute
pose) and on \epichands (leading to better generalization in the wild).

\begin{table}[t]
    \centering
    \caption{\textbf{Role of auxiliary supervision.} We consider grasp and mask supervision from both \epicb \& \ego to train WildHands and show results in zero-shot generalization settings. Both grasp \& mask supervision lead to improvements in 3D \& 2D metrics, with hand masks providing larger gain compared to grasp labels. Even though auxiliary supervision is on Epic/Ego4D, it leads to improvements in all settings, \ie benefits from training on broad data extend beyond datasets with auxiliary supervision.} 
    \tablelabel{tab:loss_ablation}
    \setlength{\tabcolsep}{6pt}
    \resizebox{\linewidth}{!}{
    \begin{tabular}{l c c c c c c}
        \toprule
        & \multicolumn{2}{c}{\bf H2O} & \multicolumn{2}{c}{\bf \assemblya} & \bf \egoexo & \bf \epichands \\
        \cmidrule(lr){2-3} \cmidrule(lr){4-5} \cmidrule(lr){6-6} \cmidrule(lr){7-7}
                                                 & \bf MPJPE & \bf MRRPE & \bf MPJPE & \bf MRRPE & \bf MPJPE & \bf L2 Error\\
        \midrule
        Wildhands (no aux)           & 39.52     & 77.07     & 93.44     & 208.32     & 70.39 & 17.07      \\ %
        \midrule
        + \epicc grasp                 & 38.34     & 76.04     & 90.23     & 180.85     & 63.30 & --         \\ %
        + \epicc mask                  & 34.29     & 60.23     & 87.94     & 175.31     & 56.41 & --         \\ %
        + \epicc grasp + \epicc mask     & \bf 31.08 & \bf 49.49 & \bf 84.91 & \bf 164.90 & \bf 55.84 & --         \\ %
        \midrule
        + \ego grasp                & 41.06     & 111.47    & 86.44     & 222.23     & 69.73 & 8.22       \\
        + \ego mask                 & 38.17     & \bf 57.93 & 82.55     & \bf 145.78 & 63.43 & 7.87       \\
        + \ego grasp + \ego mask   & \bf 35.62 & 62.10     & \bf 79.08 & 148.12     & \bf 60.80 & \bf 7.20   \\
        \bottomrule
    \end{tabular}
    }
\end{table}

\begin{table}[t]
    \centering
    \caption{\textbf{Comparison of \camenc with relevant approaches}. \camenc is more effective than other methods for dealing with perspective distortion, \eg Perspective Correction~\cite{Mehta2017THREEDV}, Perspective Crop Layers (PCL~\cite{Yu2021CVPR}), or other encodings, \eg CamConv~\cite{facil2019cam}}.
    \tablelabel{tab:perspective}
    \resizebox{\linewidth}{!}{
    \setlength{\tabcolsep}{6pt}
    \begin{tabular}{l c c c c c c}
        \toprule
        & \multicolumn{2}{c}{\bf H2O} & \multicolumn{2}{c}{\bf \assemblya} & \bf \egoexo & \bf \epichands \\
        \cmidrule(lr){2-3} \cmidrule(lr){4-5} \cmidrule(lr){6-6} \cmidrule(lr){7-7} & \bf MPJPE & \bf MRRPE & \bf MPJPE & \bf MRRPE & \bf MPJPE & \bf L2 Error\\
        \midrule
        HandNet + & & & & & & \\
        \quad CamConv & 36.86 & 67.62 & 96.72 & 180.73 & 60.69 & 17.35 \\
        \quad Perspective Corr. & 39.95 & 159.13 & 59.10 & 637.32 & 67.45 & 28.68 \\
        \quad PCL~\cite{Yu2021CVPR} & 36.82 & 158.88 & {\bf45.18} & 483.92 & 63.65 & 28.21 \\
        \quad \camenc (WildHands)    & {\bf31.08} & {\bf49.49} & 84.91 & {\bf164.90} & {\bf55.84} & {\bf11.05} \\
        \bottomrule
    \end{tabular}
    }
\end{table}

\boldparagraph{Role of auxiliary supervision}
We extract hand masks \& grasp labels from \epicb \& \ego and show their
benefits in~\Tableref{tab:loss_ablation} in zero-shot evaluation settings. Mask
supervision leads to gains of 8.5\% in MPJPE, 21.5\% in MRRPE and 55.5\% in 2D
pose. Grasp labels improve MPJPE by 2.5\%, MRRPE by 7.3\% and 2D pose by 4.3\%.
While both sources of supervision are effective, hand masks lead to larger
gains. Combining both mask and grasp supervision leads to further
improvements in both 3D \& 2D poses across most settings. Moreover,
auxiliary supervision on in-the-wild data also aids performance on lab datasets, suggesting that generalization gains from training on broad data are
not dataset specific.

\begin{table}[t]
    \centering
    \caption{\textbf{\camenc Design Choices}. We study the impact of different
    design choices of \camenc on \name: adding \camenc with the input instead
    of latent features (w/ input), removing intrinsics from \camenc (no intrx),
    dense variant of \camenc from~\cite{Prakash2023Ambiguity}. \name uses
    sparse variant of \camenc. We observe that all variants of \camenc provide
    significant benefits compared to the model without \camenc and the sparse
    variant performs the best.}
    \tablelabel{tab:design_choices}
    \resizebox{\linewidth}{!}{
    \setlength{\tabcolsep}{6pt}
    \begin{tabular}{l c c c c c c}
        \toprule
        & \multicolumn{2}{c}{\bf H2O} & \multicolumn{2}{c}{\bf \assemblya} & \bf \egoexo & \bf \epichands \\
        \cmidrule(lr){2-3} \cmidrule(lr){4-5} \cmidrule(lr){6-6} \cmidrule(lr){7-7}
                                                 & \bf MPJPE & \bf MRRPE & \bf MPJPE & \bf MRRPE & \bf MPJPE & \bf L2 Error\\
        \midrule
        no \camenc                     & 38.06     & 141.06    & 109.88    & 317.49 & 89.72  & 31.62 \\
        \camenc w/ input               & 45.51     & 80.96     & 94.45     & 252.34 & 93.56  & 17.30      \\
        \camenc no intrx               & 36.97     & 61.98     & 92.12     & 246.45 & 60.80  & 11.63      \\
        \camenc dense                  & 36.86     & 80.54     & 95.34     & 201.33 & 69.11  & 11.24 \\
        \camenc sparse                 & \bf 31.08     & \bf 49.49     & \bf 84.91     & \bf 55.84  & \bf 55.84  & \bf 11.05 \\
        \bottomrule
    \end{tabular}
    }
\end{table}

\begin{table}[t]
    \caption{\textbf{Effect of scaling up data.} Training on more datasets
    leads to consistent improvements in models performance on held out
    datasets.}
    \tablelabel{tab:dataset_ablation}
    \centering
    \setlength{\tabcolsep}{6pt}
    \resizebox{\linewidth}{!}{
    \begin{tabular}{l cccc}
        \toprule
        & \multicolumn{2}{c}{\bf H2O} & \bf \egoexo & \bf \epichands \\
        \cmidrule(lr){2-3} \cmidrule(lr){4-4} \cmidrule(lr){5-5}
                                                 & \bf MPJPE & \bf MRRPE & \bf MPJPE & \bf L2 Error\\
        \midrule
        \arctic                         & 47.30 & 75.17 & 87.71 & 17.07 \\
        \arctic + \assemblya            & 39.52 & 77.07 & 70.39 & 11.05 \\
        \arctic + Assembly + \ego (aux) & \bf 35.62 & \bf 62.10 & \bf 60.80 & \bf 7.20 \\
        \bottomrule
    \end{tabular}}
\end{table}

\boldparagraph{Comparison of \camenc with relevant approaches} In~\Tableref{tab:perspective}, we find \camenc to be more effective than other methods for dealing with perspective distortion, \eg Perspective Correction~\cite{Mehta2017THREEDV}, Perspective Crop Layers (PCL~\cite{Yu2021CVPR}), or different forms of positional encoding, \eg CamConv~\cite{facil2019cam}.

\boldparagraph{Impact on transformer models}. We investigate if our insights are useful to transformer models as well, \ie if \camenc helps on top of positional encodings already used in transformers and if auxiliary supervision leads to better generalization for large capacity models. For this, we implement these components in HandOccNet~\cite{Park2022CVPR} \& HaMeR~\cite{pavlakos2023reconstructing} and train these models in our settings. From the results in~\Tableref{tab:kpe_transformer}, we see consistent gains across all settings.

\boldparagraph{\camenc design choice} We ablate different variants of \camenc
in~\Tableref{tab:design_choices}: adding \camenc with the input instead of
latent features (w/ input), removing intrinsics from \camenc (no intrx) and
dense variant of \camenc from~\cite{Prakash2023Ambiguity}. Note that the sparse
variant performs the best, so we use sparse \camenc in \name. 

\boldparagraph{Intrinsics during training} Intrinsics may not always be
available in in-the-wild data used to derive auxiliary supervision. To study this
setting, we consider in-the-wild \ego data since it contains images from
multiple cameras, and do not assume access to intrinsics. In this case, we
replace the \camenc with a sinusoidal positional encoding of normalized image
coordinates \wrt center. The \ego results in~\Tableref{tab:loss_ablation}
follow this setting and we observe that auxiliary supervision from \ego
provides benefits even in the absence of camera information.

\boldparagraph{Scaling up training data} We ablate variants of \name trained
with \arctic, \arctic + \assembly, \arctic + \ego and \arctic + \assembly +
\ego in zero-shot settings on H2O, \egoexo, and \epichands. We use 3D
supervision on \arctic \& \assembly and auxiliary supervision (hand masks,
grasp labels) on \ego. \Tableref{tab:dataset_ablation} shows consistent
improvements in 3D and 2D metrics from both \assembly and \ego datasets,
suggesting that further scaling can improve performance further.

\begin{figure*}[t]
    \begin{tabular}{c c}
        \tile{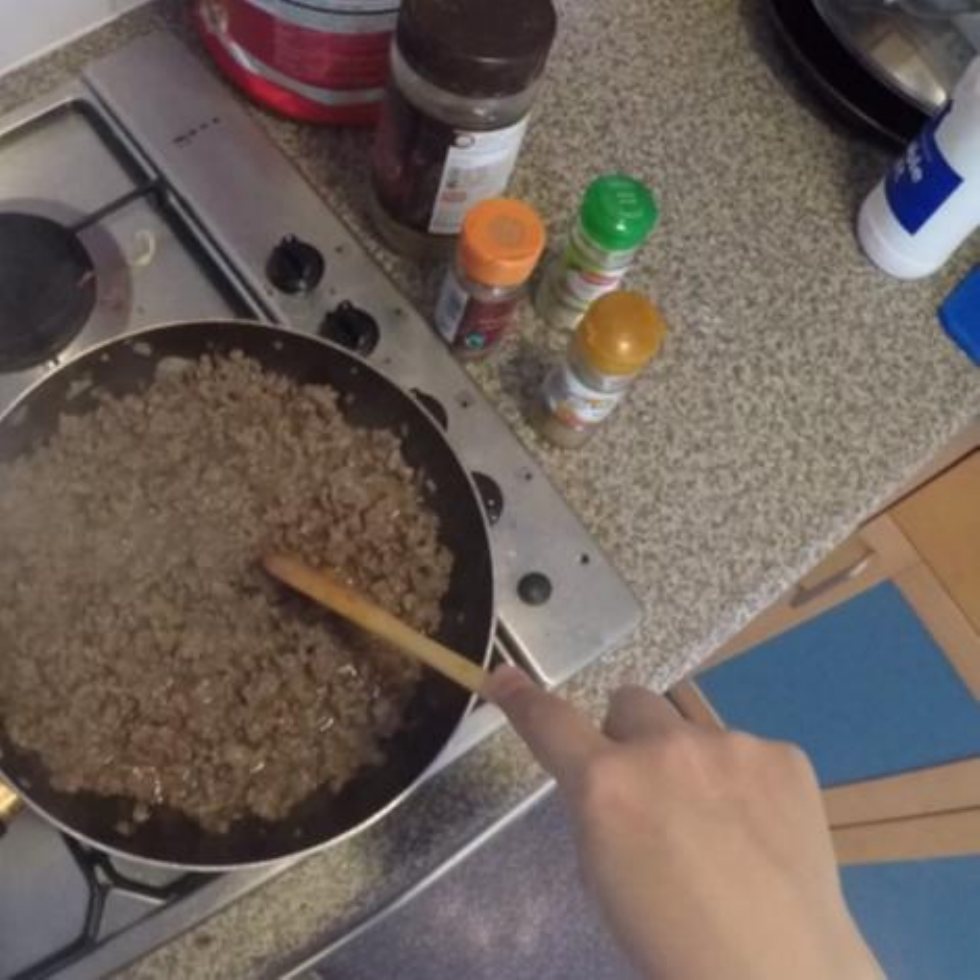}{fm}{FrankMocap}{wh}{WildHands}{0 0 0 0}
        &
        \tile{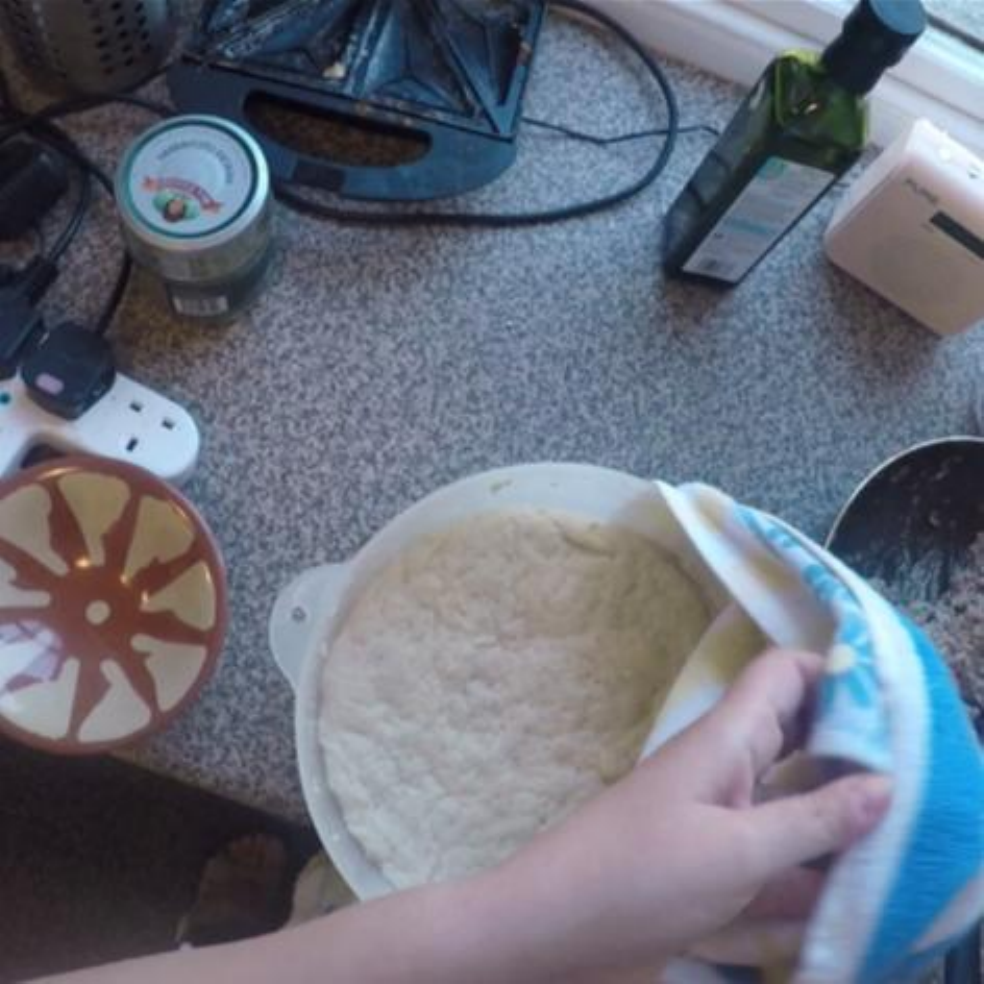}{fm}{FrankMocap}{wh}{WildHands}{0 0 0 0} \\
        \tile{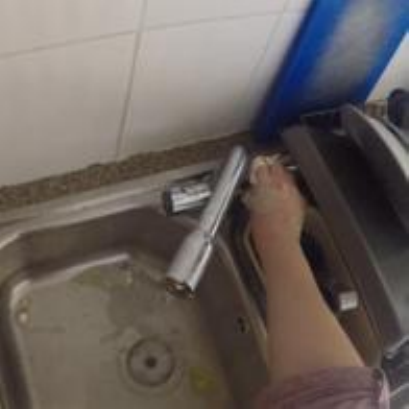}{hm}{HaMeR}{wh}{WildHands}{0 0 0 0}
        &
        \tile{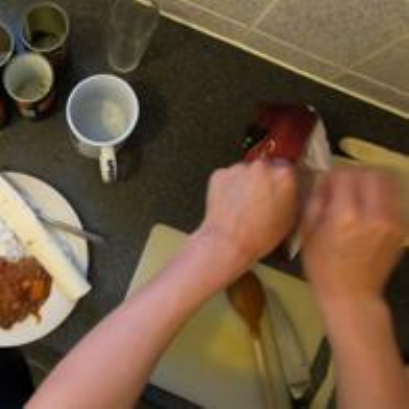}{hm}{HaMeR}{wh}{WildHands}{0 0 0 0} \\
        \tile{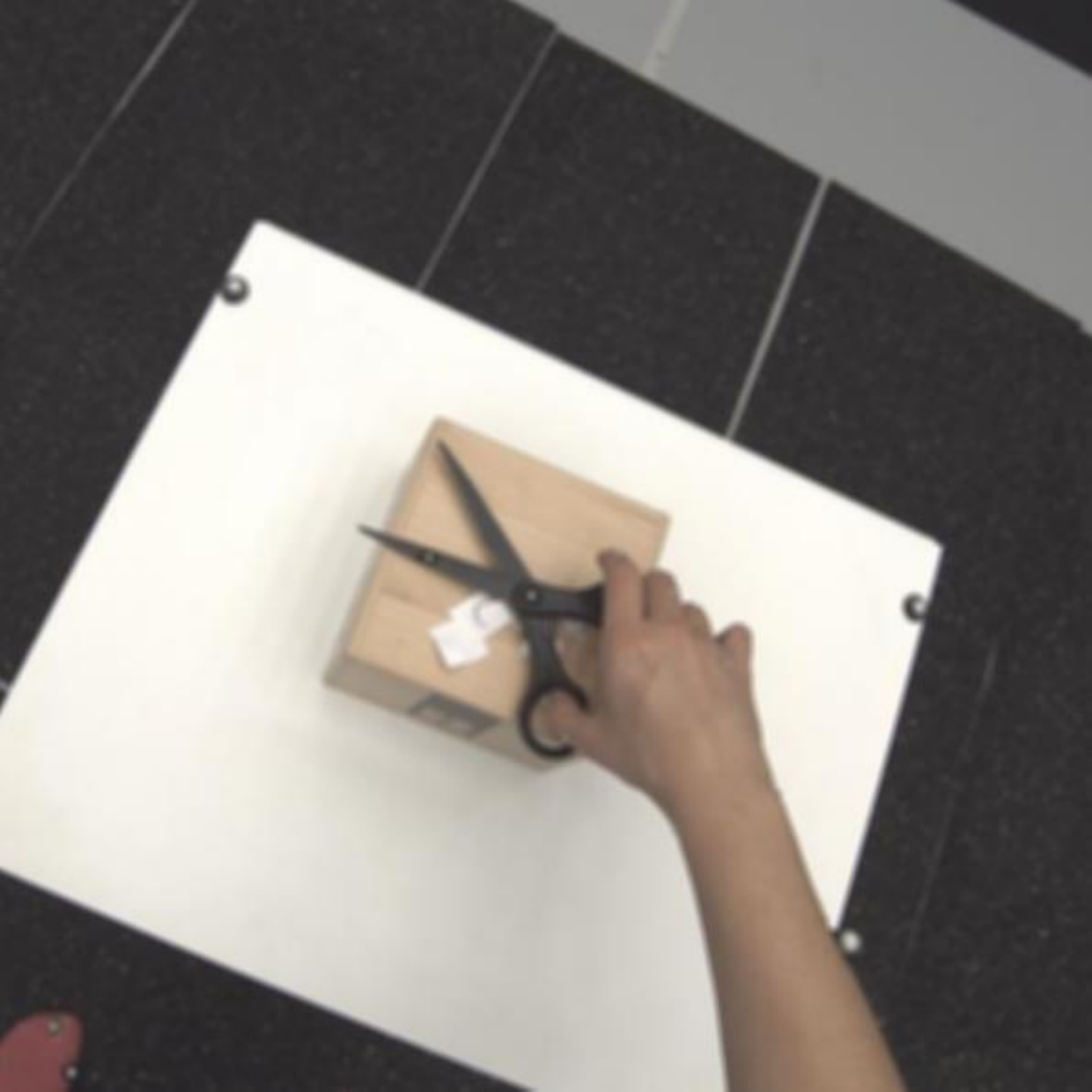}{arctic}{ArcticNet}{wh}{WildHands}{0 0 0 0}
        &
        \tile{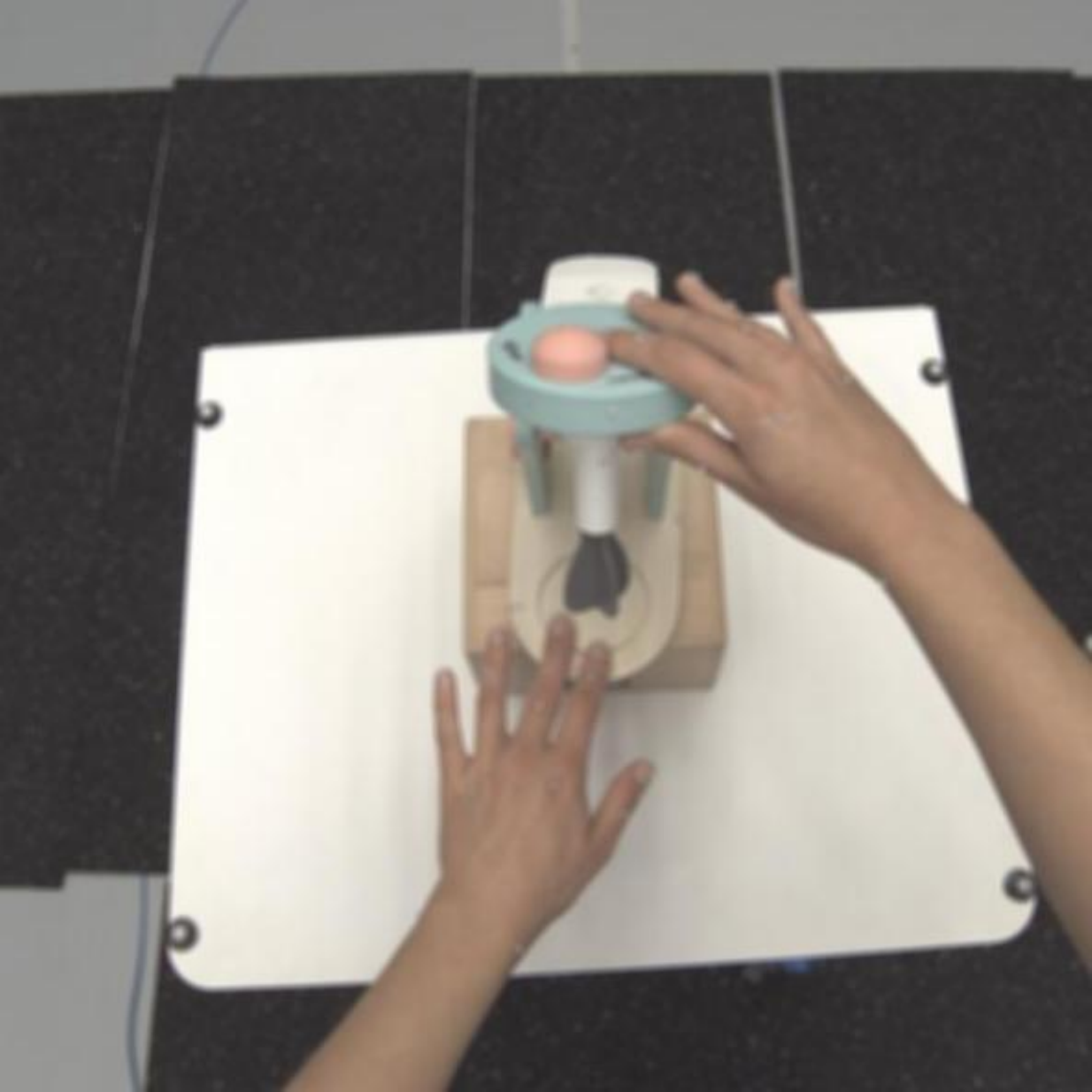}{arctic}{ArcticNet}{wh}{WildHands}{0 0 0 0} \\
    \end{tabular}
    \caption{\textbf{Visualizations}. We show projection of the predicted hand in the
    image \& rendering of the hand mesh from 2 more views.
    \name predicts better hand poses from a single image than
    FrankMocap~\cite{rong2020frankmocap}, HaMeR~\cite{Fan2023CVPR} and ArcticNet~\cite{Fan2023CVPR} in
    challenging egocentric scenarios involving occlusions and perspective distortion.}
    \figlabel{visor_vis}
\end{figure*}

\subsection{System-level Evaluation}
\setlength{\intextsep}{0pt}%
\begin{wrapfigure}[12]{r}{0.4\textwidth}
    \centering
    \captionof{table}{{\bf Leaderboard results}. \name leads the 3D hand pose on the egocentric split of \arctic leaderboard (as of July 13, 2024).}
    \tablelabel{tab:leaderboard}
    \resizebox{\linewidth}{!}{
    \begin{tabular}{l c c}
        \toprule
        \bf Method & \bf MPJPE & \bf MRRPE \\
        \midrule
        ArcticNet-SF & 19.18 & 28.31 \\
        ArcticOccNet & 19.77 & 29.75 \\
        DIGIT-HRNet & 16.74 & 25.49 \\
        HMR-ResNet50 & 20.32 & 32.32 \\
        JointTransformer & 16.33 & 26.07 \\
        \name & \textbf{15.72} & \textbf{23.88} \\
        \bottomrule
    \end{tabular}}
\end{wrapfigure}

While all of our earlier experiments are conducted in controlled settings, we also present a system-level comparison to other past methods, specifically to methods submitted to the ARCTIC leaderboard (as of July 13, 2024), and with the publicly released models of FrankMocap~\cite{rong2020frankmocap} and HaMeR~\cite{pavlakos2023reconstructing}.

\boldparagraph{ARCTIC Leaderboard}
Our method achieves the best 3D hand pose on the egocentric split, compared to recent state-of-the-art convolutional (\eg ArcticNet-SF, DIGIT-HRNet, HMR-ResNet50) and transformer (\eg JointTransformer) models (as of July 13, 2024). However, it is not possible to do a detailed comparison since most of these models are not public.

\begin{table}[t]
    \caption{\textbf{Systems comparison}. We evaluate against publicly released
    models: FrankMocap~\cite{rong2020frankmocap} (a popular method for 3D hand
    pose estimation), and HaMeR~\cite{pavlakos2023reconstructing}. FrankMocap uses a ResNet-50
    backbone and is trained on 6 lab datasets. HaMeR uses a ViT-H~\cite{dosovitskiy2020image}
    backbone and is trained on 7 lab + 3 in-the-wild + HInt datasets across nearly 3M
    frames. WildHands model uses a ResNet-50 backbone and is trained on 3
    datasets. WildHands outperforms FrankMocap across all metrics and HaMeR on 3 of 6 metrics while being 10$\times$ smaller \& trained on 5$\times$ less data. We expect scaling up the backbone and datasets
    used to train \name can lead to even stronger performance.}
    \tablelabel{system}
    \centering
    \resizebox{\linewidth}{!}{
    \setlength{\tabcolsep}{3pt}
    \begin{tabular}{l c c c c c c}
        \toprule
        & \multicolumn{2}{c}{\bf H2O} & \multicolumn{2}{c}{\bf \assemblya} & \bf \egoexo & \bf \epichands \\
        \cmidrule(lr){2-3} \cmidrule(lr){4-5} \cmidrule(lr){6-6} \cmidrule(lr){7-7} & \bf MPJPE & \bf MRRPE & \bf MPJPE & \bf MRRPE & \bf MPJPE & \bf L2 Error\\
        \midrule
        FrankMocap~\cite{rong2020frankmocap} (ResNet-50, 6 lab)      & 58.51     & -         & 97.59     & -         & 175.91        & 13.33       \\
        HaMeR~\cite{pavlakos2023reconstructing} (ViT-H, 7 lab+3 wild+HInt)   & \bf 23.82     & 147.87    & \bf 45.49     & 334.52    & 116.46        & \bf 4.56 \\
        \name (ResNet-50, 2 lab + 1 wild)  & 31.08     & \bf 49.49     & 80.40     & \bf 148.12     & \bf 55.84        & 7.20       \\
        \bottomrule
    \end{tabular}
    }
\end{table}

\boldparagraph{Comparison with FrankMocap~\cite{rong2020frankmocap} and HaMeR~\cite{pavlakos2023reconstructing}} We show results with the publicly released models in~\tableref{system}. Note that HaMeR uses a ViT-H backbone which is much larger and more performant than the ResNet50 backbone used in \name. WildHands outperforms FrankMocap across all metrics and HaMeR on 3 of 6 metrics while being 10$\times$ smaller \& trained on 5$\times$ less data.

\subsection{Visualizations}
We show qualitative comparisons of the hand pose, predicted by \name, with
FrankMocap on \epichands (\Figref{visor_vis}\,a) and ArcticNet-SF on ARCTIC
(\Figref{visor_vis}\,b). Looking at the projection of the mesh in the camera
view and rendering of the mesh from additional views, we observe that \name is
able to predict hand pose better in images involving occlusion and interaction,
\eg fingers are curled around the object in contact (\Figref{visor_vis}) for
our model but this is not the case for FrankMocap. We observe similar trends in
ARCTIC (\Figref{visor_vis}\,b) where our model predicts better hands
in contact scenarios. More results in supplementary.

\boldparagraph{Failure Cases} We observe that images in which the fingers are barely visible, \eg when kneading a dough in top row (\Figref{visor_failure}), or containing extreme poses, \eg grasps in bottom row (\Figref{visor_failure}), are quite challenging for all models.

\begin{figure*}[t]
    \begin{tabular}{c c}
        \tile{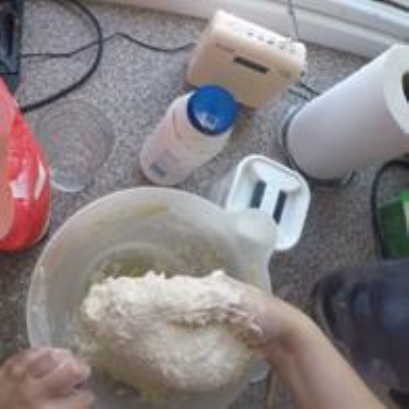}{fm}{FrankMocap}{wh}{WildHands}{0 0 0 0}
        \tile{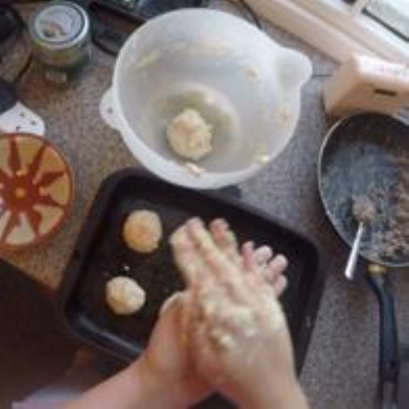}{fm}{FrankMocap}{wh}{WildHands}{0 0 0 0} \\
        \tile{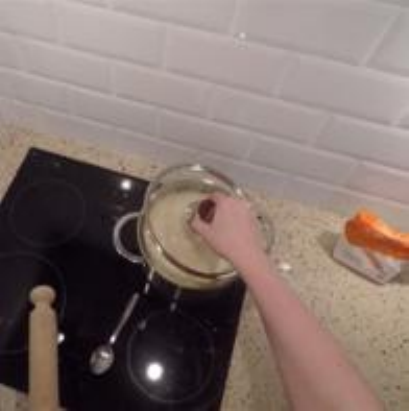}{hm}{HaMeR}{wh}{WildHands}{0 0 0 0}
        \tile{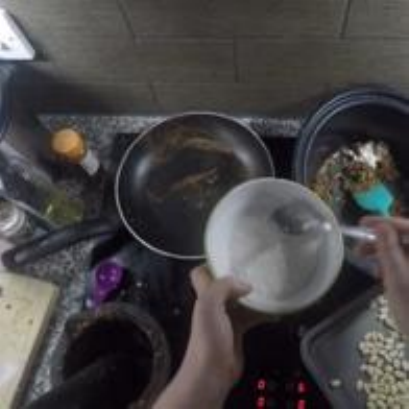}{hm}{HaMeR}{wh}{WildHands}{0 0 0 0} \\
    \end{tabular}
    \caption{\textbf{Failure cases}. We observe that images with (top) barely visible fingers, \eg kneading dough or (bottom) extreme grasp poses are challenging for all models.}
    \figlabel{visor_failure}
\end{figure*}

\boldparagraph{Limitations}
The \camenc encoding requires camera intrinsics to be known, which may not be available in certain scenarios. However, in several in-the-wild images, the metadata often contains camera information. Also, we currently set the weights for different loss terms as hyperparameters which may not be ideal since the sources of supervision are quite different leading to different scales in loss values. It could be useful to use a learned weighing scheme, \eg uncertainty-based loss weighting~\cite{brazil2023omni3d,Hu2021ICCV,Kendall2018CVPR}.

\section{Conclusion}
We present \name, a system that adapts best practices from the literature: using crops as input, intrinsics-aware positional encoding, auxiliary sources of supervision and multi-dataset training, for robust prediction of 3D hand poses on egocentric images in the wild. Experiments on both lab datasets and in-the-wild settings show the effectiveness of \name. As future direction, \name could be used to scale up learning robot policies from human interactions. %

\boldparagraph{Acknowledgements} We thank Arjun Gupta, Shaowei Liu, Anand Bhattad \& Kashyap Chitta for feedback on the draft, and David Forsyth for useful discussion. This material is based upon work supported by NSF (IIS2007035), NASA (80NSSC21K1030), DARPA (Machine Common Sense program), Amazon Research Award, NVIDIA Academic Hardware Grant, and the NCSA Delta System (supported by NSF OCI 2005572 and the State of Illinois).

\bibliographystyle{splncs04}
\bibliography{biblioShort, biblioLong, refs, refs_ambiguity}

\begin{thebibliography}{10}
\providecommand{\url}[1]{\texttt{#1}}
\providecommand{\urlprefix}{URL }
\providecommand{\doi}[1]{https://doi.org/#1}

\bibitem{Ballan2012ECCV}
Ballan, L., Taneja, A., Gall, J., Gool, L.V., Pollefeys, M.: Motion capture of hands in action using discriminative salient points. In: Proceedings of the European Conference on Computer Vision (ECCV) (2012)

\bibitem{brazil2023omni3d}
Brazil, G., Kumar, A., Straub, J., Ravi, N., Johnson, J., Gkioxari, G.: Omni3d: A large benchmark and model for 3d object detection in the wild. In: Proceedings of the IEEE Conference on Computer Vision and Pattern Recognition (CVPR). pp. 13154--13164 (2023)

\bibitem{cao2021reconstructing}
Cao, Z., Radosavovic, I., Kanazawa, A., Malik, J.: Reconstructing hand-object interactions in the wild. In: Proceedings of the IEEE International Conference on Computer Vision (ICCV) (2021)

\bibitem{Chao2021CVPR}
Chao, Y., Yang, W., Xiang, Y., Molchanov, P., Handa, A., Tremblay, J., Narang, Y.S., Wyk, K.V., Iqbal, U., Birchfield, S., Kautz, J., Fox, D.: Dexycb: {A} benchmark for capturing hand grasping of objects. In: Proceedings of the IEEE Conference on Computer Vision and Pattern Recognition (CVPR) (2021)

\bibitem{chen2019learning}
Chen, Z., Zhang, H.: Learning implicit fields for generative shape modeling. In: Proceedings of the IEEE Conference on Computer Vision and Pattern Recognition (CVPR) (2019)

\bibitem{cheng2023towards}
Cheng, T., Shan, D., Hassen, A.S., Higgins, R.E.L., Fouhey, D.: Towards a richer 2d understanding of hands at scale. In: Advances in Neural Information Processing Systems (NeurIPS) (2023)

\bibitem{Damen2018ScalingEV}
Damen, D., Doughty, H., Farinella, G.M., Fidler, S., Furnari, A., Kazakos, E., Moltisanti, D., Munro, J., Perrett, T., Price, W., Wray, M.: Scaling egocentric vision: The epic-kitchens dataset. Proceedings of the European Conference on Computer Vision (ECCV)  (2018)

\bibitem{damen2020collection}
Damen, D., Doughty, H., Farinella, G.M., Fidler, S., Furnari, A., Kazakos, E., Moltisanti, D., Munro, J., Perrett, T., Price, W., Wray, M.: The epic-kitchens dataset: Collection, challenges and baselines. IEEE Transactions on Pattern Analysis and Machine Intelligence (TPAMI)  (2020)

\bibitem{damen2018scaling}
Damen, D., Doughty, H., Farinella, G.M., Fidler, S., Furnari, A., Kazakos, E., Moltisanti, D., Munro, J., Perrett, T., Price, W., et~al.: Scaling egocentric vision: The epic-kitchens dataset. In: Proceedings of the European Conference on Computer Vision (ECCV) (2018)

\bibitem{darkhalil2022visor}
Darkhalil, A., Shan, D., Zhu, B., Ma, J., Kar, A., Higgins, R., Fidler, S., Fouhey, D., Damen, D.: Epic-kitchens visor benchmark: Video segmentations and object relations. In: NeurIPS Track on Datasets and Benchmarks (2022)

\bibitem{dosovitskiy2020image}
Dosovitskiy, A., Beyer, L., Kolesnikov, A., Weissenborn, D., Zhai, X., Unterthiner, T., Dehghani, M., Minderer, M., Heigold, G., Gelly, S., et~al.: An image is worth 16x16 words: Transformers for image recognition at scale. arXiv preprint arXiv:2010.11929  (2020)

\bibitem{facil2019cam}
Facil, J.M., Ummenhofer, B., Zhou, H., Montesano, L., Brox, T., Civera, J.: Cam-convs: Camera-aware multi-scale convolutions for single-view depth. In: Proceedings of the IEEE Conference on Computer Vision and Pattern Recognition (CVPR). pp. 11826--11835 (2019)

\bibitem{Fan20213DV}
Fan, Z., Spurr, A., Kocabas, M., Tang, S., Black, M.J., Hilliges, O.: Learning to disambiguate strongly interacting hands via probabilistic per-pixel part segmentation. In: Proceedings of the International Conference on 3D Vision (3DV) (2021)

\bibitem{Fan2023CVPR}
Fan, Z., Taheri, O., Tzionas, D., Kocabas, M., Kaufmann, M., Black, M.J., Hilliges, O.: {ARCTIC:} {A} dataset for dexterous bimanual hand-object manipulation. In: Proceedings of the IEEE Conference on Computer Vision and Pattern Recognition (CVPR) (2023)

\bibitem{freeman1995orientation}
Freeman, W.T., Roth, M.: Orientation histograms for hand gesture recognition. In: International workshop on automatic face and gesture recognition. vol.~12, pp. 296--301. Citeseer (1995)

\bibitem{garcia2018first}
Garcia-Hernando, G., Yuan, S., Baek, S., Kim, T.K.: First-person hand action benchmark with rgb-d videos and 3d hand pose annotations. In: Proceedings of the IEEE Conference on Computer Vision and Pattern Recognition (CVPR) (2018)

\bibitem{grauman2022ego4d}
Grauman, K., Westbury, A., Byrne, E., Chavis, Z., Furnari, A., Girdhar, R., Hamburger, J., Jiang, H., Liu, M., Liu, X., et~al.: Ego4d: Around the world in 3,000 hours of egocentric video. In: Proceedings of the IEEE Conference on Computer Vision and Pattern Recognition (CVPR) (2022)

\bibitem{grauman2023ego}
Grauman, K., Westbury, A., Torresani, L., Kitani, K., Malik, J., Afouras, T., Ashutosh, K., Baiyya, V., Bansal, S., Boote, B., et~al.: Ego-exo4d: Understanding skilled human activity from first-and third-person perspectives. arXiv preprint arXiv:2311.18259  (2023)

\bibitem{guizilini2023towards}
Guizilini, V., Vasiljevic, I., Chen, D., Ambruș, R., Gaidon, A.: Towards zero-shot scale-aware monocular depth estimation. In: Proceedings of the IEEE International Conference on Computer Vision (ICCV) (2023)

\bibitem{guizilini2022depth}
Guizilini, V., Vasiljevic, I., Fang, J., Ambru, R., Shakhnarovich, G., Walter, M.R., Gaidon, A.: Depth field networks for generalizable multi-view scene representation. In: Proceedings of the European Conference on Computer Vision (ECCV) (2022)

\bibitem{hampali2020honnotate}
Hampali, S., Rad, M., Oberweger, M., Lepetit, V.: Honnotate: A method for 3d annotation of hand and object poses. In: Proceedings of the IEEE Conference on Computer Vision and Pattern Recognition (CVPR) (2020)

\bibitem{Hampali2022CVPR}
Hampali, S., Sarkar, S.D., Rad, M., Lepetit, V.: Keypoint transformer: Solving joint identification in challenging hands and object interactions for accurate 3d pose estimation. In: Proceedings of the IEEE Conference on Computer Vision and Pattern Recognition (CVPR) (2022)

\bibitem{hartley2003multiple}
Hartley, R., Zisserman, A.: Multiple view geometry in computer vision. Cambridge university press (2003)

\bibitem{Hasson2020LeveragingPC}
Hasson, Y., Tekin, B., Bogo, F., Laptev, I., Pollefeys, M., Schmid, C.: Leveraging photometric consistency over time for sparsely supervised hand-object reconstruction. Proceedings of the IEEE Conference on Computer Vision and Pattern Recognition (CVPR)  (2020)

\bibitem{hasson19_obman}
Hasson, Y., Varol, G., Tzionas, D., Kalevatykh, I., Black, M.J., Laptev, I., Schmid, C.: Learning joint reconstruction of hands and manipulated objects. In: Proceedings of the IEEE Conference on Computer Vision and Pattern Recognition (CVPR) (2019)

\bibitem{He2017ICCV}
He, K., Gkioxari, G., Doll{\'{a}}r, P., Girshick, R.B.: Mask {R-CNN}. In: Proceedings of the IEEE International Conference on Computer Vision (ICCV) (2017)

\bibitem{He2016CVPR}
He, K., Zhang, X., Ren, S., Sun, J.: Deep residual learning for image recognition. In: Proceedings of the IEEE Conference on Computer Vision and Pattern Recognition (CVPR) (2016)

\bibitem{heap1996towards}
Heap, T., Hogg, D.: Towards 3d hand tracking using a deformable model. In: Proceedings of the Second International Conference on Automatic Face and Gesture Recognition. pp. 140--145. Ieee (1996)

\bibitem{Hu2021ICCV}
Hu, A., Murez, Z., Mohan, N., Dudas, S., Hawke, J., Badrinarayanan, V., Cipolla, R., Kendall, A.: {FIERY:} future instance prediction in bird's-eye view from surround monocular cameras. In: Proceedings of the IEEE International Conference on Computer Vision (ICCV) (2021)

\bibitem{Ioffe2015ICML}
Ioffe, S., Szegedy, C.: Batch normalization: Accelerating deep network training by reducing internal covariate shift. In: Bach, F.R., Blei, D.M. (eds.) Proceedings of the International Conference on Machine Learning (ICML) (2015)

\bibitem{Ivashechkin2023ARXIV}
Ivashechkin, M., Mendez, O., Bowden, R.: Denoising diffusion for 3d hand pose estimation from images. arXiv  \textbf{2308.09523} (2023)

\bibitem{Jiang2023CVPR}
Jiang, C., Xiao, Y., Wu, C., Zhang, M., Zheng, J., Cao, Z., Zhou, J.T.: A2j-transformer: Anchor-to-joint transformer network for 3d interacting hand pose estimation from a single {RGB} image. In: Proceedings of the IEEE Conference on Computer Vision and Pattern Recognition (CVPR) (2023)

\bibitem{JiangR2023CVPRa}
Jiang, Z., Rahmani, H., Black, S., Williams, B.M.: A probabilistic attention model with occlusion-aware texture regression for 3d hand reconstruction from a single {RGB} image. In: Proceedings of the IEEE Conference on Computer Vision and Pattern Recognition (CVPR) (2023)

\bibitem{kanazawa2018end}
Kanazawa, A., Black, M.J., Jacobs, D.W., Malik, J.: End-to-end recovery of human shape and pose. In: Proceedings of the IEEE Conference on Computer Vision and Pattern Recognition (CVPR) (2018)

\bibitem{kanazawa2018learning}
Kanazawa, A., Tulsiani, S., Efros, A.A., Malik, J.: Learning category-specific mesh reconstruction from image collections. In: Proceedings of the European Conference on Computer Vision (ECCV) (2018)

\bibitem{karunratanakul2020grasping}
Karunratanakul, K., Yang, J., Zhang, Y., Black, M.J., Muandet, K., Tang, S.: Grasping field: Learning implicit representations for human grasps. In: Proceedings of the International Conference on 3D Vision (3DV) (2020)

\bibitem{kato2018neural}
Kato, H., Ushiku, Y., Harada, T.: Neural 3d mesh renderer. In: Proceedings of the IEEE Conference on Computer Vision and Pattern Recognition (CVPR) (2018)

\bibitem{Kendall2018CVPR}
Kendall, A., Gal, Y., Cipolla, R.: Multi-task learning using uncertainty to weigh losses for scene geometry and semantics. In: Proceedings of the IEEE Conference on Computer Vision and Pattern Recognition (CVPR) (2018)

\bibitem{Kingma2015ICLR}
Kingma, D.P., Ba, J.: Adam: {A} method for stochastic optimization. In: Bengio, Y., LeCun, Y. (eds.) Proceedings of the International Conference on Learning Representations (ICLR) (2015)

\bibitem{kwon2021h2o}
Kwon, T., Tekin, B., St{\"u}hmer, J., Bogo, F., Pollefeys, M.: H2o: Two hands manipulating objects for first person interaction recognition. In: Proceedings of the IEEE International Conference on Computer Vision (ICCV) (2021)

\bibitem{Lin2017CVPR}
Lin, T., Doll{\'{a}}r, P., Girshick, R.B., He, K., Hariharan, B., Belongie, S.J.: Feature pyramid networks for object detection. In: Proceedings of the IEEE Conference on Computer Vision and Pattern Recognition (CVPR) (2017)

\bibitem{Liu2019ICCV}
Liu, S., Chen, W., Li, T., Li, H.: Soft rasterizer: {A} differentiable renderer for image-based 3d reasoning. In: Proceedings of the IEEE International Conference on Computer Vision (ICCV) (2019)

\bibitem{liu2020general}
Liu, S., Li, T., Chen, W., Li, H.: A general differentiable mesh renderer for image-based 3d reasoning. IEEE Transactions on Pattern Analysis and Machine Intelligence (TPAMI)  (2020)

\bibitem{Liu2022CVPR}
Liu, Y., Liu, Y., Jiang, C., Lyu, K., Wan, W., Shen, H., Liang, B., Fu, Z., Wang, H., Yi, L.: {HOI4D:} {A} 4d egocentric dataset for category-level human-object interaction. In: Proceedings of the IEEE Conference on Computer Vision and Pattern Recognition (CVPR) (2022)

\bibitem{Mehta2017THREEDV}
Mehta, D., Rhodin, H., Casas, D., Fua, P., Sotnychenko, O., Xu, W., Theobalt, C.: Monocular 3d human pose estimation in the wild using improved {CNN} supervision. In: Proceedings of the International Conference on 3D Vision (3DV) (2017)

\bibitem{Mildenhall2020ECCV}
Mildenhall, B., Srinivasan, P.P., Tancik, M., Barron, J.T., Ramamoorthi, R., Ng, R.: Nerf: Representing scenes as neural radiance fields for view synthesis. In: Proceedings of the European Conference on Computer Vision (ECCV) (2020)

\bibitem{miyato2023gta}
Miyato, T., Jaeger, B., Welling, M., Geiger, A.: {GTA}: A geometry-aware attention mechanism for multi-view transformers. arXiv  (2023)

\bibitem{Moon2020ECCV}
Moon, G., Yu, S., Wen, H., Shiratori, T., Lee, K.M.: Interhand2.6m: {A} dataset and baseline for 3d interacting hand pose estimation from a single {RGB} image. In: Proceedings of the European Conference on Computer Vision (ECCV) (2020)

\bibitem{Nair2010ICML}
Nair, V., Hinton, G.E.: Rectified linear units improve restricted boltzmann machines. In: Proceedings of the International Conference on Machine Learning (ICML) (2010)

\bibitem{ohkawa2023assemblyhands}
Ohkawa, T., He, K., Sener, F., Hodan, T., Tran, L., Keskin, C.: Assemblyhands: Towards egocentric activity understanding via 3d hand pose estimation. In: Proceedings of the IEEE Conference on Computer Vision and Pattern Recognition (CVPR). pp. 12999--13008 (2023)

\bibitem{Park2022CVPR}
Park, J., Oh, Y., Moon, G., Choi, H., Lee, K.M.: Handoccnet: Occlusion-robust 3d hand mesh estimation network. In: Proceedings of the IEEE Conference on Computer Vision and Pattern Recognition (CVPR) (2022)

\bibitem{pavlakos2023reconstructing}
Pavlakos, G., Shan, D., Radosavovic, I., Kanazawa, A., Fouhey, D., Malik, J.: Reconstructing hands in 3d with transformers. arXiv preprint arXiv:2312.05251  (2023)

\bibitem{Potamias_2023_CVPR}
Potamias, R.A., Ploumpis, S., Moschoglou, S., Triantafyllou, V., Zafeiriou, S.: Handy: Towards a high fidelity 3d hand shape and appearance model. In: Proceedings of the IEEE Conference on Computer Vision and Pattern Recognition (CVPR). pp. 4670--4680 (June 2023)

\bibitem{Prakash2023Ambiguity}
Prakash, A., Gupta, A., Gupta, S.: Mitigating perspective distortion-induced shape ambiguity in image crops. arXiv  \textbf{2312.06594} (2023)

\bibitem{ravi2020pytorch3d}
Ravi, N., Reizenstein, J., Novotny, D., Gordon, T., Lo, W.Y., Johnson, J., Gkioxari, G.: Accelerating 3d deep learning with pytorch3d. arXiv:2007.08501  (2020)

\bibitem{rehg1994visual}
Rehg, J.M., Kanade, T.: Visual tracking of high dof articulated structures: an application to human hand tracking. In: Proceedings of the European Conference on Computer Vision (ECCV) (1994)

\bibitem{rogez20143d}
Rogez, G., Khademi, M., Supan{\v{c}}i{\v{c}}~III, J., Montiel, J.M.M., Ramanan, D.: 3d hand pose detection in egocentric rgb-d images. In: Proceedings of the European Conference on Computer Vision (ECCV) (2014)

\bibitem{romero2017embodied}
Romero, J., Tzionas, D., Black, M.J.: Embodied hands: Modeling and capturing hands and bodies together. ACM Transactions on Graphics (ToG)  (2017)

\bibitem{rong2020frankmocap}
Rong, Y., Shiratori, T., Joo, H.: Frankmocap: Fast monocular {3D} hand and body motion capture by regression and integration. Proceedings of the IEEE International Conference on Computer Vision Workshops (ICCV Workshops)  (2021)

\bibitem{Sener2022CVPR}
Sener, F., Chatterjee, D., Shelepov, D., He, K., Singhania, D., Wang, R., Yao, A.: Assembly101: {A} large-scale multi-view video dataset for understanding procedural activities. In: Proceedings of the IEEE Conference on Computer Vision and Pattern Recognition (CVPR) (2022)

\bibitem{shan2020understanding}
Shan, D., Geng, J., Shu, M., Fouhey, D.F.: Understanding human hands in contact at internet scale. In: Proceedings of the IEEE Conference on Computer Vision and Pattern Recognition (CVPR) (2020)

\bibitem{sharp2015accurate}
Sharp, T., Keskin, C., Robertson, D., Taylor, J., Shotton, J., Kim, D., Rhemann, C., Leichter, I., Vinnikov, A., Wei, Y., et~al.: Accurate, robust, and flexible real-time hand tracking. In: Proceedings of the 33rd annual ACM conference on human factors in computing systems. pp. 3633--3642 (2015)

\bibitem{Simon2017CVPR}
Simon, T., Joo, H., Matthews, I.A., Sheikh, Y.: Hand keypoint detection in single images using multiview bootstrapping. In: Proceedings of the IEEE Conference on Computer Vision and Pattern Recognition (CVPR) (2017)

\bibitem{sridhar2016real}
Sridhar, S., Mueller, F., Zollh{\"o}fer, M., Casas, D., Oulasvirta, A., Theobalt, C.: Real-time joint tracking of a hand manipulating an object from rgb-d input. In: Proceedings of the European Conference on Computer Vision (ECCV) (2016)

\bibitem{Sridhar2013ICCV}
Sridhar, S., Oulasvirta, A., Theobalt, C.: Interactive markerless articulated hand motion tracking using {RGB} and depth data. In: Proceedings of the IEEE International Conference on Computer Vision (ICCV) (2013)

\bibitem{Sun2015CVPR}
Sun, X., Wei, Y., Liang, S., Tang, X., Sun, J.: Cascaded hand pose regression. In: Proceedings of the IEEE Conference on Computer Vision and Pattern Recognition (CVPR) (2015)

\bibitem{taheri2020grab}
Taheri, O., Ghorbani, N., Black, M.J., Tzionas, D.: {GRAB}: A dataset of whole-body human grasping of objects. In: Proceedings of the European Conference on Computer Vision (ECCV) (2020)

\bibitem{tompson2014real}
Tompson, J., Stein, M., Lecun, Y., Perlin, K.: Real-time continuous pose recovery of human hands using convolutional networks. ACM Transactions on Graphics (ToG)  \textbf{33}(5),  1--10 (2014)

\bibitem{tulsiani2017multi}
Tulsiani, S., Zhou, T., Efros, A.A., Malik, J.: Multi-view supervision for single-view reconstruction via differentiable ray consistency. In: Proceedings of the IEEE Conference on Computer Vision and Pattern Recognition (CVPR). pp. 2626--2634 (2017)

\bibitem{tzionas20153d}
Tzionas, D., Gall, J.: 3d object reconstruction from hand-object interactions. In: Proceedings of the IEEE International Conference on Computer Vision (ICCV) (2015)

\bibitem{Vaswani2017AttentionIA}
Vaswani, A., Shazeer, N.M., Parmar, N., Uszkoreit, J., Jones, L., Gomez, A.N., Kaiser, L., Polosukhin, I.: Attention is all you need. Advances in Neural Information Processing Systems (NeurIPS)  (2017)

\bibitem{Wan2016ECCV}
Wan, C., Yao, A., Gool, L.V.: Hand pose estimation from local surface normals. In: Proceedings of the European Conference on Computer Vision (ECCV) (2016)

\bibitem{Yang2022CVPR}
Yang, L., Li, K., Zhan, X., Wu, F., Xu, A., Liu, L., Lu, C.: Oakink: {A} large-scale knowledge repository for understanding hand-object interaction. In: Proceedings of the IEEE Conference on Computer Vision and Pattern Recognition (CVPR) (2022)

\bibitem{ye2022ihoi}
Ye, Y., Gupta, A., Tulsiani, S.: What's in your hands? {3D} reconstruction of generic objects in hands. In: Proceedings of the IEEE Conference on Computer Vision and Pattern Recognition (CVPR) (2022)

\bibitem{yifan2022input}
Yifan, W., Doersch, C., Arandjelovi{\'c}, R., Carreira, J., Zisserman, A.: Input-level inductive biases for 3d reconstruction. In: Proceedings of the IEEE Conference on Computer Vision and Pattern Recognition (CVPR) (2022)

\bibitem{Yu2021CVPR}
Yu, F., Salzmann, M., Fua, P., Rhodin, H.: Pcls: Geometry-aware neural reconstruction of 3d pose with perspective crop layers. In: Proceedings of the IEEE Conference on Computer Vision and Pattern Recognition (CVPR) (2021)

\bibitem{zhang2019end}
Zhang, X., Li, Q., Mo, H., Zhang, W., Zheng, W.: End-to-end hand mesh recovery from a monocular rgb image. In: ICCV (2019)

\bibitem{zimmermann2017learning}
Zimmermann, C., Brox, T.: Learning to estimate 3d hand pose from single rgb images. In: Proceedings of the IEEE International Conference on Computer Vision (ICCV) (2017)

\bibitem{Zimmermann2019ICCV}
Zimmermann, C., Ceylan, D., Yang, J., Russell, B.C., Argus, M.J., Brox, T.: Freihand: {A} dataset for markerless capture of hand pose and shape from single {RGB} images. In: Proceedings of the IEEE International Conference on Computer Vision (ICCV) (2019)

\end{thebibliography}
\end{document}